%% file: main.tex
\documentclass{article}

\usepackage{arxiv}

\usepackage[utf8]{inputenc} 
\usepackage[T1]{fontenc}    
\usepackage{hyperref}       
\usepackage{url}            
\usepackage{booktabs}       
\usepackage{amsfonts}       
\usepackage{nicefrac}       
\usepackage{microtype}      
\usepackage{lipsum}
\usepackage{graphicx}
\usepackage[utf8]{inputenc} 
\usepackage[T1]{fontenc}    
\usepackage{hyperref}       
\usepackage{url}            
\usepackage{booktabs}       
\usepackage{amsfonts}       
\usepackage{nicefrac}       
\usepackage{microtype}      
\usepackage{xcolor}         
\usepackage{graphicx}
\usepackage{subfigure}
\usepackage{amsmath}
\usepackage{amsthm}
\usepackage{amssymb}
\usepackage{adjustbox}
\usepackage{wrapfig}
\usepackage[normalem]{ulem}
\usepackage{mathrsfs}
\usepackage{multirow}
\usepackage{svg}

\usepackage[linesnumbered,ruled,vlined]{algorithm2e}

\DeclareMathOperator*{\softmax}{softmax}
\DeclareMathOperator*{\arcface}{arcface}
\graphicspath{ {./figures/} }

\title{Condensed Prototype Replay for Class Incremental Learning}

\author{
 Jiangtao Kong \\
 Department of Computer Science \\
  College of William and Mary \\
  \texttt{jkong01@email.wm.edu} \\
   \And
 Zhenyu Zong \\
   Department of Computer Science \\
  College of William and Mary \\
  \texttt{zzong@email.wm.edu} \\
  \And
  Tianyi Zhou \\
  Department of  Computer Science \\
  University of Maryland, College Park \\
  \texttt{tianyi@umd.edu} \\
  \And
  Huajie Shao \\
 Department of Computer Science \\
  College of William and Mary \\
  \texttt{hshao@wm.edu} \\
}

\begin{document}
\maketitle
\begin{abstract}
\input{abstract}
\end{abstract}


\input{intro}
\input{model}
\input{evaluation}

\input{relatedwork}

\input{conclude}

\bibliographystyle{abbrv}
\bibliography{reference.bib}

\clearpage
\appendix

\input{appendix}

\end{document}

%% file: abstract.tex
Incremental learning (IL) suffers from catastrophic forgetting of old tasks when learning new tasks. This can be addressed by replaying previous tasks' data stored in a memory, which however is usually prone to size limits and privacy leakage. Recent studies store only class centroids as prototypes and augment them with Gaussian noises to create synthetic data for replay. However, they cannot effectively avoid class interference near their margins that leads to forgetting. Moreover, the injected noises distort the rich structure between real data and prototypes, hence even detrimental to IL.      
In this paper, we propose YONO that You Only Need to replay One condensed prototype per class, which for the first time can even outperform memory-costly exemplar-replay methods. To this end, we develop a novel prototype learning method that (1) searches for more representative prototypes in high-density regions by an attentional mean-shift algorithm and (2) moves samples in each class to their prototype to form a compact cluster distant from other classes. Thereby, the class margins are maximized, which effectively reduces interference causing future forgetting. In addition, we extend YONO to YONO+, which 
creates synthetic replay data by random sampling in the neighborhood of each prototype in the representation space. We show that the synthetic data can further improve YONO. Extensive experiments on IL benchmarks demonstrate the advantages of YONO/YONO+ over existing IL methods in terms of both accuracy and forgetting.

%% file: intro.tex
\section{Introduction}\label{sec:intro}
Catastrophic forgetting~\cite{mccloskey1989catastrophic} refers to deep neural networks forget the acquired knowledge from the previous tasks disastrously while learning the current task. This is in sharp contrast to humans who are able to incrementally learn new knowledge from the ever-changing world. To bridge the gap between artificial intelligence and human intelligence, incremental learning (IL)~\cite{wu2019large,gepperth2016incremental,douillard2022dytox,xie2022general} has emerged as a new paradigm to enable AI systems to  continuously learn from new data over time.


In the past few years, a variety of methods~\cite{roady2020stream,cong2020gan,wang2021ordisco,xue2022meta} have been proposed to alleviate catastrophic forgetting in IL. In this work, we are interested in a very challenging scenario, called class-incremental learning (CIL)~\cite{zhu2021prototype,zhou2022model,zhu2022self}. CIL aims to identify all the previously learned classes with no task identifier available at the inference time. Unfortunately, CIL often suffers from catastrophic forgetting because of the overlapping representations between the previous tasks and the current one in the feature space~\cite{zhu2021prototype}. To deal with this issue, many prior studies adopt \textit{exemplar-based approaches} to preserve some old class samples in a memory buffer. These methods, however, suffer from memory limitations and privacy issues. Thus, some works propose \textit{non-exemplar-based methods}~\cite{li2017learning, yu2020semantic, lopez2017gradient, mallya2018packnet} that incrementally learn new tasks without storing raw samples in a memory. Most of these methods mainly focus on building large network structures~\cite{mallya2018packnet}, or designing regularization loss~\cite{li2017learning} to mitigate catastrophic forgetting, but they do not perform well in CIL.


\begin{wrapfigure}[18]{r}{0.54\textwidth}
  \begin{center}
    \includegraphics[width=0.5\textwidth]{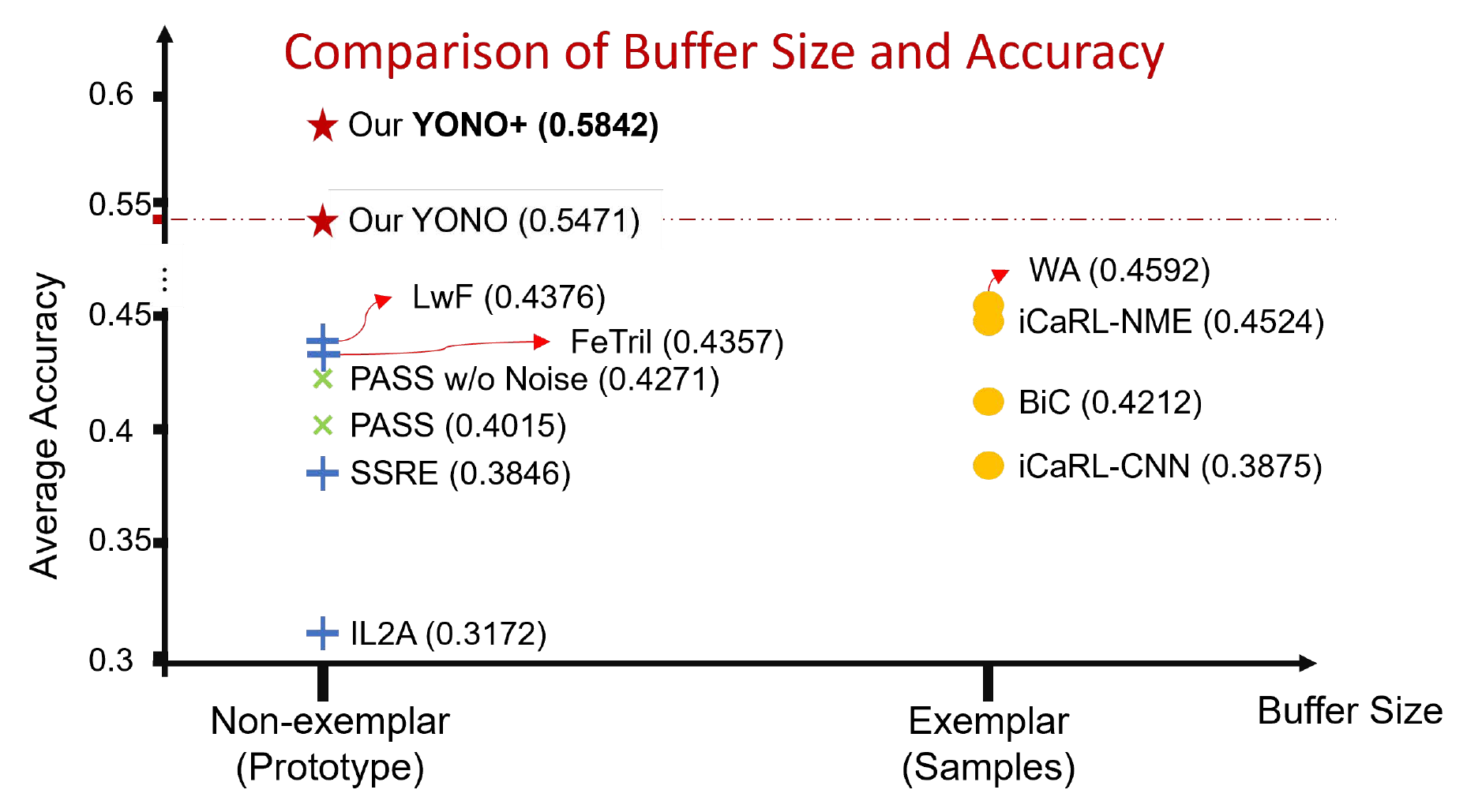}
  \end{center}
  \caption{Comparison of buffer size and accuracy for different methods on TinyImageNet under zero-base and 10 phrases setting. YONO that only stores and replays one prototype for each class can achieve higher accuracy than the baselines.}
  \label{fig:motivation}
\end{wrapfigure}
\vspace{-0.05in}

Recently, few studies~\cite{zhu2021prototype, zhu2022self}, such as PASS and SSRE, propose to store one prototype (\textit{class mean}) for each old class and then use augmented prototypes to train a model. Surprisingly, we find that the PASS~\cite{zhu2021prototype} using prototype augmentation via Gaussian noise will degrade the prediction accuracy compared to that without prototype augmentation, as illustrated in Fig.~\ref{fig:motivation}. This is because class mean may not represent the centroid of different representations in a high-dimensional space such that the recovered representations of old classes may overlap with similar classes. It thus motivates us to optimize the learning of prototype for each class in CIL.

In this work, we develop YONO, a new non-exemplar model that \textit{only needs to store and replay one class-representative prototype}. The key challenge lies in how to find a more representative prototype for each class so that these prototypes can be distant from each other. To address this challenge, we propose a new attentional mean-shift method to dynamically aggregate the representations of samples in each class into a prototype in a high-density region. Then we only replay one prototype for each old class without using synthetic data when training a model. To further improve the accuracy of YONO, we extend it to YONO+ that generates synthetic data from stored prototypes. Accordingly, we develop a novel approach that combines a high-dimensional space rotation matrix and Gaussian distribution to create synthetic data of old classes from stored prototypes. Extensive experiments are carried out to evaluate the performance of our methods on multiple benchmarks. Experimental results demonstrate that both YONO and YONO+ can significantly outperform the baselines in terms of accuracy and average forgetting. Moreover, replaying synthetic data can further improve the performance of YONO.

\textbf{Our contributions} are four-fold: 1) we propose a novel non-exemplar model, called YONO, that can achieve good performance by only replaying one stored prototype for each class without using synthetic data, 2) to our best knowledge, we are the first to explore the prototype optimization in IL, 3) we extend YONO to YONO+, which develops a new data synthesis technique that creates high-quality data from stored prototypes, 4) the evaluation results demonstrate the superiority of our methods over the non-exemplar baselines in terms of accuracy and average forgetting.

%% file: model.tex
\section{Proposed Method}
\label{sec:model}
In this section, we first elaborate on the proposed YONO that adopts attentional mean-shift method to compute a representative prototype for each class. In order to further improve the prediction accuracy, we extend YONO to develop a YONO+ with synthetic data generated from prototypes.

\begin{figure*}[!tb]
\centering
\includegraphics[width=0.98\linewidth]{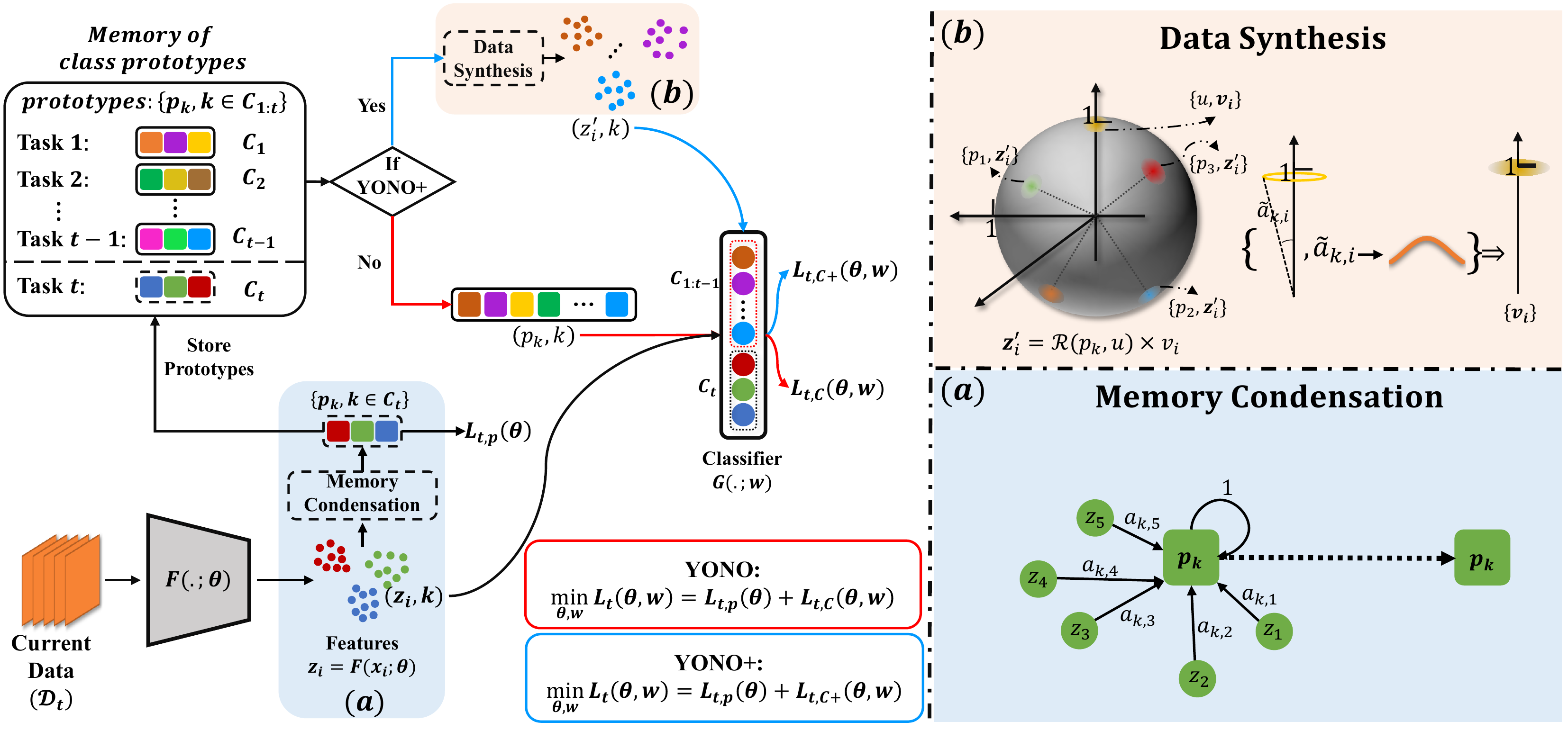}
\caption{Framework of the proposed YONO and YONO+. YONO only needs to replay one stored prototype for each class while YONO+ is trained on synthetic data generated from stored prototypes. (a)~Memory condensation learns a compact prototype for each class inspired by mean shift algorithm. (b) Data synthesis aims to generate the representations of old classes from stored prototypes using a $m$-dimensional space rotation matrix and Gaussian distribution.
}
\label{fig:pipeline}
\vspace{-0.2cm}
\end{figure*}

\noindent \textbf{Problem Description.} 
By learning from a sequence tasks each associated with a subset of classes $C_t$ and a training set of $n_t$ examples drawn from these classes, i.e., $\mathcal{D}_t\triangleq\{x_i, y_i\}_{i=1}^{n_t}$ with $y_i\in C_t$, class-incremental learning (CIL) aims to train a model $f(x;[\theta,w])\triangleq G(F(x; \theta); w)$ that predicts probabilities of all previous classes $C_{1:t}\triangleq\bigcup_{i=1}^{t} C_i$ for any given example $x$. The model is composed of a feature extractor $F(\cdot; \theta)$ producing compact representations and a classifier $G(\cdot; w)$. Given $x$, the probabilities over all classes $C_{1:t}$ are predicted as $\softmax(wF(x; \theta))$.

\subsection{YONO}\label{sec:RMB_method}
In the following, we introduce YONO, which performs two main steps in each iteration: (i) prototype learning for memory condensation that computes a prototype per class; and (ii) new task learning with prototype replay. 

For (i), we propose a novel attentional mean-shift method to compute a prototype for each class as a mode searching of the class distribution. By training the representation for prototype-based classification, we are able to concentrate most samples to their class prototypes and keep each class a compact cluster in the representation space. This strategy significantly mitigates inter-class interference, which is a primary reason for forgetting. 

For (ii), when learning a new task, we augment its training set with previous tasks' class prototypes from the memory. In YONO, replaying only the prototypes of previous classes suffice to retain their classes' features and mitigate catastrophic forgetting. 

\subsubsection{Prototype Learning for Memory Condensation}
\label{sec:aggregation}

When learning task-$t$ defined on a set of classes $C_t$, for each class $k\in C_t$, we construct a graph of the class's sample representations $z_i=F(x_i;\theta)$ that connects class-$k$'s prototype $p_k$ and apply graph attention~\cite{velivckovic2017graph} to move $p_k$ towards a high-density region in the representation space. We achieve this by mean-shift of the prototype: in each step, we move $p_k$ towards a weighted average over all samples belonging to class-$k$ (their normalized representations in specific) and normalize the new $p_k$, i.e.,
\begin{equation}\label{equ:pk}
    p_k\leftarrow (1-\lambda)p_k+\lambda\sum_{i\in [n_t]:y_i=k} a_{k,i}\cdot\frac{z_i}{\|z_i\|_2},\quad p_k\leftarrow\frac{p_k}{\|p_k\|_2},
\end{equation}
where $\lambda$ controls the step size of the mean-shift and $n_t$ is the size of training set for task-$t$. Unlike the original mean-shift algorithm, the weights $a_{k,i}$ are determined by learnable dot-product attention between each sample $z_i$ and the prototype $p_k$ in the representation space, i.e.,
\begin{equation}\label{equ:attention}
a_k\triangleq\softmax(\bar a_k), \quad \bar a_k\triangleq [\bar a_{k,1},\cdots, \bar a_{k,n_t}], \quad \bar a_{k,i}=c(z_i,p_k)\triangleq\frac{\langle z_i, p_k\rangle}{\Vert z_i\Vert_2 \cdot \Vert p_k\Vert_2}. 
\end{equation}
In practice, when the number of samples $n_t$ is large, we can apply a mini-batch version of Eq.~\eqref{equ:pk} for multiple steps, where $i\in [n_t]$ is replaced by $i\in B$ ($B$ is a mini-batch of samples). We then store the prototype of each class in the memory, which will be used to train the model together with learned tasks' prototypes and a new task's data. 

\subsubsection{New Task Learning with Prototype Replay}
In YONO, we train the representation model $F(\cdot;\theta)$ to produce $z_i=F(x_i;\theta)$ for each sample $x_i$ to be close to its class prototype and distant from other classes' prototypes. We achieve this by minimizing the Arcface~\cite{deng2019arcface} loss for task-$t$, i.e.,
\begin{align}\label{equ:prototype_loss}
L_{t,P}(\theta)\triangleq\frac{1}{n_t}\sum_{k\in C_t}\sum_{i\in [n_t]:y_i=k} \arcface(z_i, p,k).
\end{align}
The $\text{arcface}(\cdot,\cdot,\cdot)$ loss is defined by
\begin{equation}\label{equ:arcface}
\arcface(z,p,k) = -\log\frac{\exp(\cos[c^{-1}(z,p_k)+\delta]/\tau)}{\exp(\cos[c^{-1}(z,p_k)+\delta]/\tau)+\sum_{l\in C_{1:t},l\neq k}\exp(c(z,p_l)/\tau)},
\end{equation}
where $c^{-1}(z,p_k)\triangleq\arccos(c(z,p_k))$ denotes the angle between $z$ and $p_k$, 
$\tau$ is a temperature parameter and $\delta$ is a margin penalty to restrict the angle $c^{-1}(z_i,p_k)$ from being too large. 
Hence, samples belonging to each class are enforced to concentrate on their prototype and form a compact cluster distant from other classes in the representation space, which effectively reduces the harmful interference between classes that leads to future catastrophic forgetting. 

Meanwhile, in order to mitigate the catastrophic forgetting of previous tasks' classes $C_{1:t-1}$, YONO replays the stored class prototypes while training with the current task's data, which means we augment the training set for task-$t$ with prototypes from previous classes $C_{1:t-1}$.
The corresponding training objective for classification on $C_{1:t}$ is to minimize the negative log-likelihood on all the task-$t$'s data and prototypes for all the previous $t-1$ tasks, i.e., 
\begin{align}\label{equ:classifier_loss}
L_{t,C}(\theta,w)\triangleq \frac{1}{n_t}\sum_{k\in C_t}\sum_{i\in [n_t]:y_i=k}\arcface(z_i, w, k) + \frac{1}{|C_{1:{t-1}}|}\sum_{k\in C_{1:{t-1}}}\arcface(p_k, w, k).
\end{align}
Hence, the training objective $L_t(\theta,w)$ of YONO at task-$t$ combines the prototype-learning loss for task-$t$ in Eq.~\ref{equ:prototype_loss} and the prototype-replay augmented loss in Eq.~\eqref{equ:classifier_loss}, i.e.,
\begin{equation}\label{equ:yono}
{\rm YONO:}~~\min_{\theta,w} L_t(\theta,w)=L_{t,P}(\theta)+L_{t,C}(\theta,w).
\end{equation}
In summary, $L_{t,P}(\theta)$ mainly focuses on moving the current task's samples to their associated class prototype in the representation space so the prototypes retain most information of the task. On the other hand, $L_{t,C}(\theta,w)$ trains the representation model's parameter $\theta$ and the classifier layer(s) $w$ in an end-to-end manner on an augmented dataset composed of both the current task's data and the prototypes so the model can learn new tasks without suffering from forgetting previous tasks. 

\subsection{YONO+} \label{sec:recovery}
Although prototype-only replay in YONO is highly effective in mitigating catastrophic forgetting, it might be insufficient to cover all useful information of the whole distribution for each class without replay on different instances. Hence, we propose an extension YONO+ with the replay of synthetic data generated from the prototypes in the memory. 

\subsubsection{Data Synthesis from Prototypes}


By using $\bar a_{k,i}$ computed in Eq.~\eqref{equ:attention} for each learned (and can not be accessed) sample $z_i$, we are able to synthesize a data point $z'_i$ that has a similar angular distance to the prototype $p_k$ as $z_i$ for replay. This leads to YONO+ whose replay of each previous class is conducted on multiple synthetic data points instead of a single prototype.  

In particular, \textbf{we firstly derive a rotation matrix} $\mathcal{R}(p_k,\boldsymbol{u})$ that can recover $p_k$ from a unit vector $\boldsymbol{u}=[1,0,\cdots, 0]$ on an unit $m$-sphere, i.e., $p_k=\mathcal{R}(p_k,\boldsymbol{u})\times \boldsymbol{u}$. 
To synthesize a sample $z'_i$ of class-$k$ as a proxy to $z_i$ (a previously learned sample of class-$k$), \textbf{we then randomly draw $\boldsymbol{v_i}$} in the vicinity of $\boldsymbol{u}$,  
i.e.,
\begin{equation}\label{equ:distribution}
    \boldsymbol{v_i}=[\tilde a_{k,i}, \epsilon_2, \cdots, \epsilon_m],\quad \tilde a_{k,i}\sim T\mathcal N(\mu,\sigma,\mu-\kappa\sigma, \mu+\kappa\sigma),
\end{equation}
where $T\mathcal N(\mu,\sigma,\mu-\kappa\sigma, \mu+\kappa\sigma)$ denotes a symmetric truncated Gaussian distribution with a mean of $\mu$, a variance of $\sigma$, and a lower/upper bound of $\mu\pm \kappa\sigma$. 
To make sure that $\|\boldsymbol{v_i}\|_2=1$, we draw $\epsilon_i\sim \mathcal N(0,1)$ for $i\in\{2,\cdots,m\}$ at first and then rescale them by
$\epsilon_i\leftarrow\sqrt{\nicefrac{1-(\tilde a_{k,i}+\epsilon_1)^2}{\sum_{i=2}^m\epsilon_i^2}}\cdot \epsilon_i.$
Empirically, by choosing $\kappa=1.96$, which leads to a small variance $(1-\mu)/\kappa$ of $\tilde a_{k,i}$, the synthetic data $z'_i$ are close to their prototypes and have a similar distribution as the real data $z_i$.
Thereby, we have $\boldsymbol{u}^T\boldsymbol{v}_i=\tilde a_{k,i}$, whose distribution approximates the distribution of cosine similarity $\bar a_{k,i}$ between real sample $z_i$ and its associated class prototype $p_k$.


\textbf{Next, we create $z'_i$ from $\boldsymbol{v}_i$.} 
As $p_k=\mathcal{R}(p_k,\boldsymbol{u}) \times \boldsymbol{u}$, we can apply the same rotation matrix $\mathcal{R}(p_k,\boldsymbol{u})$ to $\boldsymbol{v_i}$ to achieve $z'_i$, i.e.,
\begin{equation}\label{equ:synthesis}
z'_i = \mathcal{R}(p_k,\boldsymbol{u})\times \boldsymbol{v_i}.
\end{equation}
By applying the same rotation, the similarity between $\boldsymbol{u}$ and $\boldsymbol{v_i}$ is preserved between $p_k$ and $z'_i$. By sampling the synthetic data point $z'_i$ for each previously removed sample $z_i$ using the above synthesis, we are able to create a dataset for all seen classes in $C_{1:t}$ that can be used in the replay. 

\subsubsection{New Task Learning with Synthetic Data Replay}
\label{sec:reliable}
When learning a new task-$t$, YONO+ also replays the synthetic dataset $\mathcal{D}'_t$ generated from all previous tasks' prototypes $p_k$, i.e.,
\begin{equation}
    \mathcal{D}'_t\triangleq \left\{(z'_i,k):k\in C_{1:t-1}, z'_i = \mathcal{R}(p_k,\boldsymbol{u})\times \boldsymbol{v_i}, \boldsymbol{v_i}=[\tilde a_{k,i}, \epsilon_2, \cdots, \epsilon_m]\right\}.
\end{equation}
The training objective for task-$t$ with the replay of previous tasks' data synthesized from the stored prototypes is 
\begin{align}\label{equ:plus_classifier_loss}
L_{t,C+}(\theta,w)\triangleq \frac{1}{n_t}\sum_{k\in C_t}\sum_{i\in [n_t]:y_i=k}\arcface(z_i, w, k) + \frac{1}{|\mathcal{D}'_{t}|}\sum_{(z,k)\in\mathcal{D}'_t}\arcface(z, w, k).
\end{align}
Hence, the training objective $L_t(\theta, w)$ of YONO+ at task-$t$ combines the prototype-learning loss for task-$t$ in Eq.~\ref{equ:prototype_loss} and the synthetic-data replay augmented loss in Eq.~\eqref{equ:plus_classifier_loss}, i.e.,
\begin{equation}\label{equ:yono+}
    {\rm YONO+:}~~\min_{\theta,w} L_t(\theta, w)=L_{t,P}(\theta)+L_{t,C+}(\theta,w).
\end{equation}

\subsection{Practical Improvement to YONO/YONO+}
\label{sec:meta}

Finally, we adopt the following techniques to further enhance the model performance.

\textbf{Knowledge Distillation.}
Following previous incremental learning methods~\cite{zhu2021prototype,hou2019learning}, we apply knowledge distillation (KD)~\cite{hou2019learning} when training $F(\cdot;\theta)$ on the current task data $x\sim\mathcal D_t$ by minimizing the difference between $F(x;\theta)$ and the representations $F(x;\theta_{t-1})$ produced by previous task model $\theta_{t-1}$, i.e.,
\begin{equation}\label{equ:kdloss}
    L_{t,KD}(\theta)\triangleq \frac{1}{n_t}\sum_{i\in [n_t]}\|F(x_i;\theta) - F(x_i;\theta_{t-1})\|_2^2.
\end{equation}
Minimizing the above KD loss aims at retaining the knowledge of the previous task's model. In YONO and YONO+, we can augment their objectives $L_t(\theta,w)$ in Eq.~\eqref{equ:yono} and Eq.~\eqref{equ:yono+} with $L_{t,KD}(\theta)$. 

\textbf{Model Interpolation.} 
In addition to KD, we apply model interpolation to retain the knowledge of the previous model $\theta_{t-1}$ and avoid overfitting to the current task. Specifically, after learning task-$t$, we update the current $\theta_t$ by the following interpolation between $\theta_{t-1}$ and $\theta_t$, i.e.,
\begin{equation}
\label{eq:theta_update}
    \theta_t\leftarrow (1-\beta)\theta_{t-1}+\beta\theta_t,
\end{equation}
where $\beta\in[0,1]$ and we set $\beta=0.6$ in experiments. Since $\theta_t$ is mainly trained on task-$t$, such simple interpolation between $\theta_{t-1}$ and $\theta_t$ leads to a more balanced performance on all tasks. 

\textbf{``Partial Freezing'' of Classifier.}
Each row $w_k$ in the classifier parameter $w$ corresponds to a class $k\in C_{1:t}$. Since the current task-$t$ mainly focuses on classes $k\in C_t$, we apply a much smaller learning rate $\eta'\ll \eta$ ($\eta$ is the learning rate for other parameters) to $w_k$ associated with previous classes to avoid significant drift of their classifier parameters, i.e.,
\begin{equation}
\label{eq:eta_classifier}
    w_k\leftarrow w_k - \eta'\nabla_{w_k}L_t(\theta, w),~~\forall k\in C_{1:t-1}
\end{equation}
We provide the complete procedure of YONO and YONO+ in Algorithm~\ref{alg:YONO}.

\SetKwComment{Comment}{/* }{ */}
\SetKwProg{Init}{init}{}{}
\SetKwInOut{Input}{input}
\SetKwInOut{Output}{output}
\SetKwInOut{Init}{initialize}
\begin{algorithm}[t]\small
\caption{YONO and YONO+}\label{alg:YONO}
\Input{Training data $\mathcal D_{1:T}$ with classes $C_{1:T}$, epochs $E$, steps $S$, iterates $R$, learning rate $\eta, \eta'$, $\delta,\beta$}
\Init{Memory $\mathcal M\leftarrow\emptyset$, $\theta, w$}
\For{$t=1 \to T$}{
\For{epoch=$1 \to E$}{
    Compute features $z_i=F(x_i;\theta)$ for $(x_i,y_i)\in \mathcal D_t$\; 
    Compute prototype $p_k$ for every class $k\in C_t$ by iterating Eq.~\eqref{equ:pk} for $R$ iterations\; 
    Save prototypes: $\mathcal M\leftarrow \mathcal M\cup \{(p_k,k):k\in C_t\}$\;
    \For{step=$1 \to S$}{
        Draw a mini-batch of data $(x_i, y_i)\sim \mathcal D_t$\;
        \eIf{YONO+}{
            Data Synthesis: create a mini-batch of data $(z'_i, k)$ of previous classes from prototypes $\{p_k:k\in C_{1:t-1}\}$ by Eq.~\eqref{equ:distribution}-\eqref{equ:synthesis}\;
            Compute loss $L_t(\theta,w)$ in Eq.~\eqref{equ:yono+} on the two mini-batches\;
        }{
            Draw a mini-batch of prototypes $(p_k, k)$ from $k\in C_{1:t-1}$\;
            Compute loss $L_t(\theta,w)$ in Eq.~\eqref{equ:yono} on the two mini-batches\;
        }
        Update feature extractor: $\theta\leftarrow\theta-\eta\nabla_\theta L_t(\theta,w)$\;
        Update classifier for $k\in C_t$: $w_k\leftarrow\theta-\eta\nabla_{w_k} L_t(\theta,w)$\;
        Update classifier for $k\in C_{1:t-1}$: $w_k\leftarrow\theta-\eta'\nabla_{w_k} L_t(\theta,w)$\;
    }
}
Model interpolation: $\theta\leftarrow (1-\beta)\theta'+\beta\theta$\;
Save current task model as $\theta'\leftarrow \theta$\;
}
\Output{Feature extractor $F(\cdot;\theta)$ and classifier $G(\cdot;w)$}
\end{algorithm}

%% file: evaluation.tex
\section{Experiment}\label{sec:experiment}
In this section, we first evaluate the performance of the proposed YONO and YONO+ on CIFAR-100~\cite{krizhevsky2009learning} and TinyImageNet~\cite{yao2015tiny}. Then we evaluate the quality of synthetic data generated from memorized prototypes. Finally, we do ablation studies to explore the impact of main components and certain hyperparameters on model performance. The detailed model configurations and hyperparameter settings are presented in Appendix~\ref{app:setup}.

\noindent \textbf{Baselines.}
We compare the proposed YONO and YONO+ with non-exemplar-based methods, including LwF~\cite{li2017learning}, PASS~\cite{zhu2021prototype}, SSRE~\cite{zhu2022self}, IL2A~\cite{zhu2021class}, and FeTrIL~\cite{petit2023fetril}. We also compare them with some exemplar-based methods, such as iCaRL~\cite{rebuffi2017icarl}, BiC~\cite{wu2019large}, and WA~\cite{zhao2020maintaining}. Following prior work~\cite{rebuffi2017icarl}, we respectively report the results of CNN predictions (i.e., iCaRL-CNN) and nearest-mean-of-exemplars classification (i.e., iCaRL-NME) for the iCaRL. We measure the performance of different methods with two commonly used metrics in IL: average accuracy~\cite{rebuffi2017icarl} and average forgetting~\cite{chaudhry2018riemannian}.


\subsection{Model Configurations and Hyper-parameter Settings}\label{app:setup}
We implement the proposed methods in PyTorch~\cite{paszke2017automatic} and run the baselines using PyCIL~\cite{zhou2021pycil}, which is a well-known toolbox for CIL. In the experiments, we train the ResNet-18~\cite{he2016deep} from scratch using the SGD~\cite{ruder2016overview} optimizer with an initial learning rate of 0.01. Then the learning rate is multiplied by 0.1 per 20 epochs. The weights for prototype learning, classification and KD loss in YONO and YONO+ are 1, 1 and 30, respectively. We train the model with batch size $256$ for 60 epochs in each task. The experimental results are averaged over three random seeds. We execute different incremental phases (i.e., 5 and 10 phases) under zero-base setting. Specifically, we evenly split the classes of each dataset into several tasks. Following prior work~\cite{zhu2021prototype}, the classes in each dataset are arranged in a fixed random order. As for the memory size of exemplar-based approaches mentioned above, we use \textit{herd election}~\cite{rebuffi2017icarl} to select exemplars of previous tasks under different settings. Specifically, we store 20 samples of each old classes following the settings in~\cite{hou2019learning, rebuffi2017icarl}.

\subsection{Evaluation Results}
\begin{figure*}[!t]
\centering
\subfigure[CIFAR-base0-phases5]{
\includegraphics[width=0.47\linewidth]{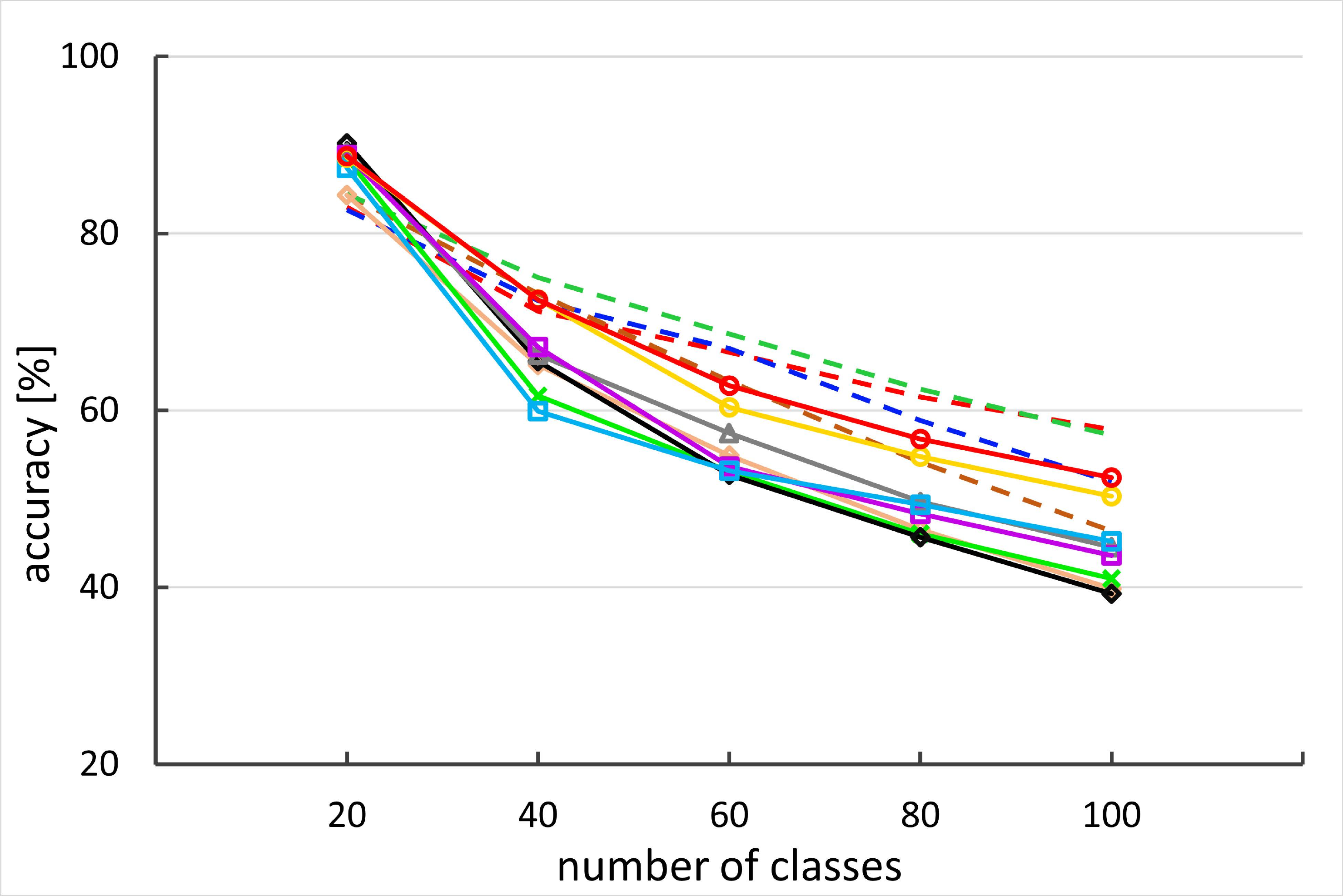}}
\label{fig:cifar-b0-5}
\subfigure[CIFAR-base0-phases10]{
\includegraphics[width=0.50\linewidth]{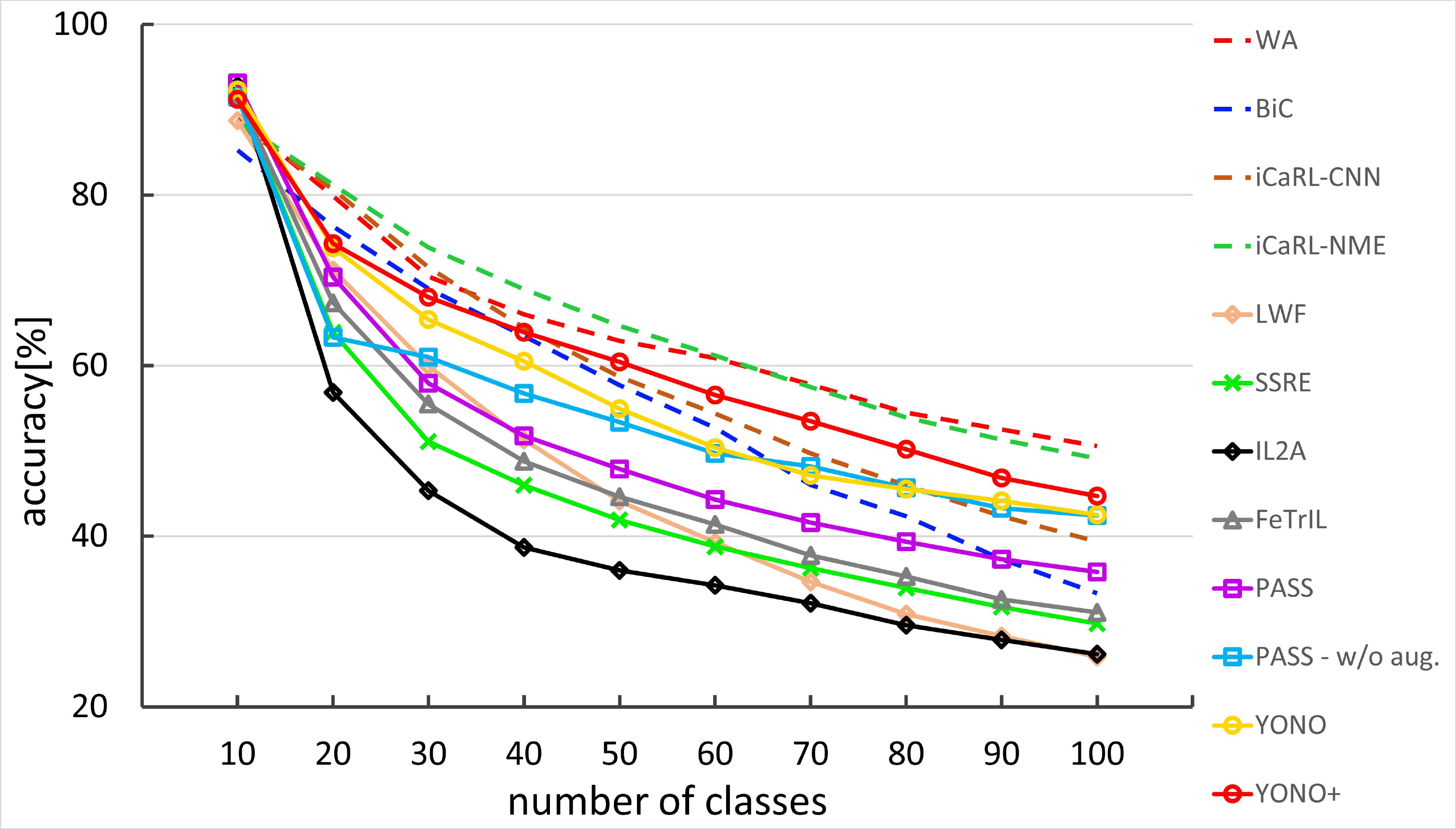}}
\label{fig:cifar-b0-10}
\subfigure[Tiny-base0-phases5]{
\includegraphics[width=0.47\linewidth]{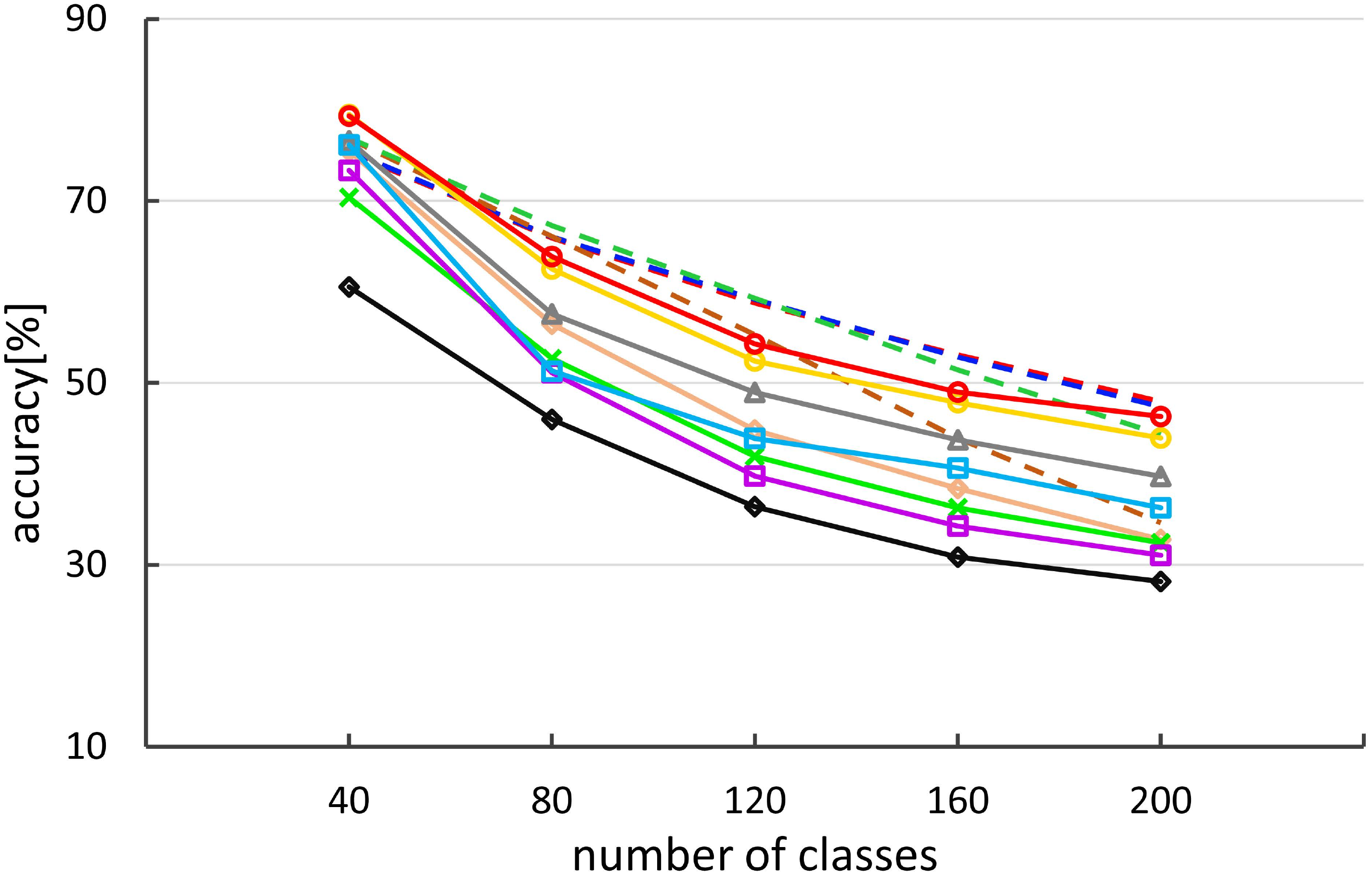}}
\label{fig:tiny-b0-5}
\subfigure[Tiny-base0-phases10]{
\includegraphics[width=0.50\linewidth]{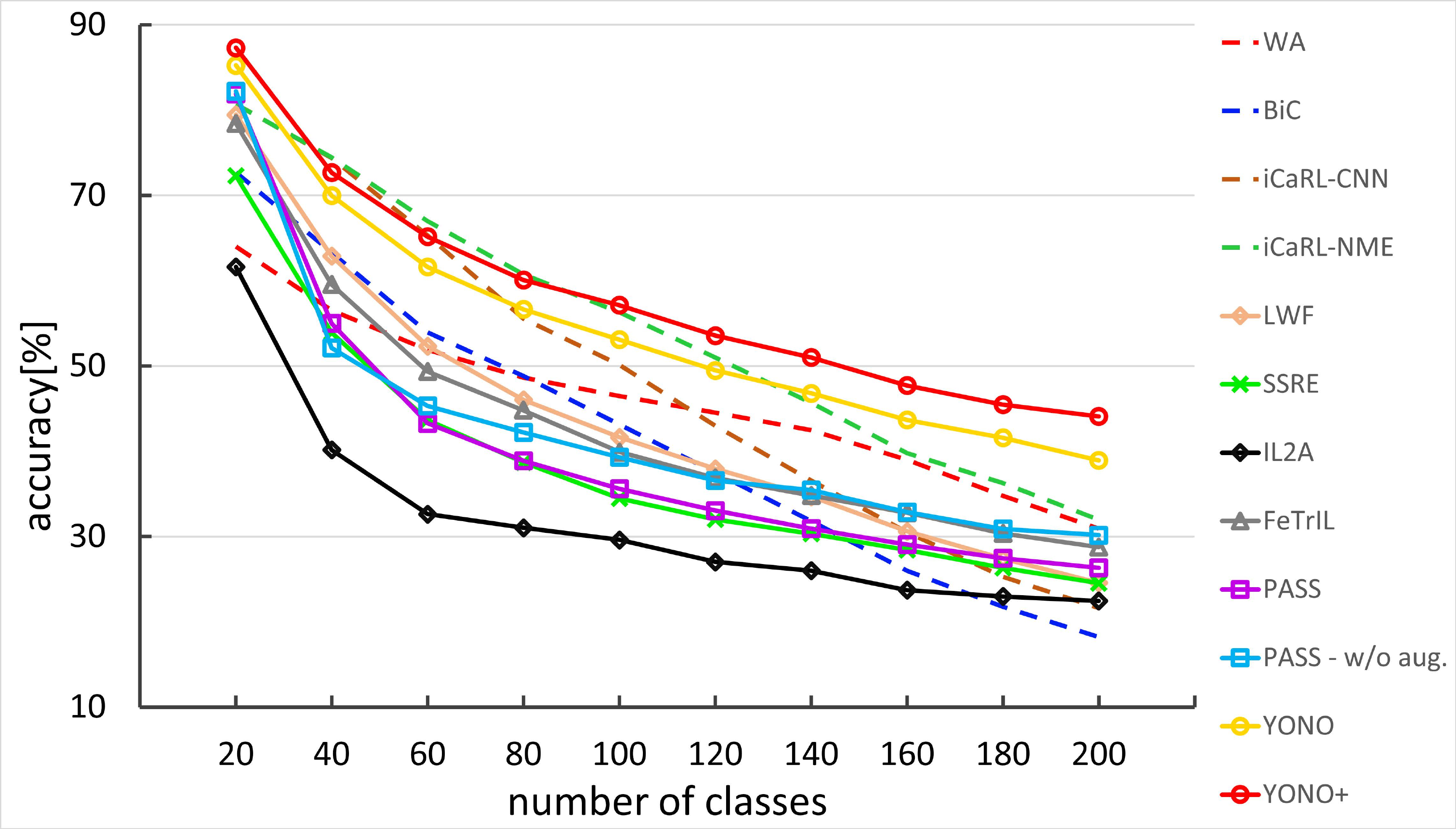}}
\label{fig:tiny-b0-10}
\caption{Accuracy comparison of different methods on CIFAR-100 and TinyImageNet under different settings. Solid lines represent non-exemplar-based approaches while the dashed lines denote exemplar-based methods.}
\label{fig:sota_cifar_tiny}
\vspace{-0.1in}
\end{figure*}

\noindent
First, we compare the proposed two methods with the baselines on CIFAR-100 and TinyImageNet under different settings: 5 phases and 10 phases. As shown in Fig.~\ref{fig:sota_cifar_tiny}, we can observe that both YONO and YONO+ outperform all the non-exemplar methods in terms of accuracy under zero-base setting. In particular, YONO can achieve comparable accuracy to YONO+ with data synthesis from stored prototypes. The reason why both YONO and YONO+ outperform the PASS is that our attentional mean-shift method can learn a compact prototype in a high-density region for each class, which reduces the inter-class interference to mitigate forgetting. We also compare our approaches with some \textit{exemplar-based models}. We can observe from Fig.~\ref{fig:sota_cifar_tiny} (d) that the proposed YONO and YONO+ outperform some exemplar-based methods on TinyImageNet.


Moreover, we present a comparison of average accuracy and forgetting for different methods, as shown in Table.~\ref{tabel:avg_forgetting}. It can be seen that the proposed YONO and YONO+ achieve higher average accuracy than other non-exemplar baselines. While SSRE and FeTrIL have lower average forgetting than our methods, their accuracy drops rapidly in the initial tasks, resulting in low accuracy in the final task, as shown in Figure~\ref{fig:sota_cifar_tiny}. In reality, the lower forgetting in SSRE and FeTrIL is attributed to the sharp drop in accuracy in the early tasks and the limited learning progress in the subsequent tasks. Moreover, the proposed methods can even outperform some exemplar-based methods in terms of average accuracy and forgetting.


\begin{table}[!t]
\caption{Average accuracy and forgetting of the proposed YONO and baselines on CIFAR-100  and TinyImageNet under different settings.``b0-10" means zero-base with 10 phases, ``b0-5" means zero-base with 5 phases. \textbf{Bold: the best among non-exemplar methods.}, {\color{red} Red: the second best among non-exemplar methods}, and {\color{blue} Blue: the best among exemplar-based methods}}
\label{tabel:avg_forgetting}
\centering
\begin{adjustbox}{width=0.95\textwidth}
\begin{tabular}{c|c|cc|cc|cc|cc}
\toprule
\multicolumn{10}{c}{Average Accuracy and Forgetting on CIFAR-100 and TinyImageNet}      \\ \hline
\multicolumn{2}{c|}{\multirow{2}{*}{Method}} &\multicolumn{2}{c|}{CIFAR-Acc [\%]$\uparrow$} &\multicolumn{2}{c|}{CIFAR-Fgt [\%]$\downarrow$} &\multicolumn{2}{c|}{Tiny-Acc [\%]$\uparrow$} &\multicolumn{2}{c}{Tiny-Fgt [\%]$\downarrow$} \\ \cline{3-10}
\multicolumn{2}{c|}{} &b0-5 &b0-10 &b0-5 &b0-10 &b0-5 &b0-10 &b0-5 &b0-10 \\\hline
\multirow{4}{*}{Exemplar} &iCaRL-CNN~\cite{rebuffi2017icarl} &64.29	&59.54	&42.91	&51.11	&51.05	& 38.75	&51.85	&59.06 \\
&iCaRL-NME~\cite{rebuffi2017icarl} & \color{blue} 69.55 & \color{blue} 65.03 & 22.82	&31.71	&56.34	&45.24	&33.41	&40.57 \\
&BiC~\cite{wu2019large} &66.57	&56.34	& \color{blue} 10.38 & \color{blue} 17.04 &  60.20 &42.12	&\color{blue}17.32	& \color{blue} 20.92 \\
&WA~\cite{zhao2020maintaining} &68.02	&64.44	&21.34	&28.91	& \color{blue} 60.21 & \color{blue}  45.92 &20.46 &32.17 \\ \hline
\hline
\multirow{8}{*}{Non-Exp} &LwF~\cite{li2017learning} &58.15	&47.43	&43.80	&51.80	&49.58	&43.76	&45.79	&54.40 \\
&SSRE~\cite{zhu2022self} &58.05	&46.58	& \textbf{15.44} & \textbf{12.13} & 46.74 & 38.47 &16.25	&19.94 \\
&IL2A~\cite{zhu2021class} &59.78	&41.96	&26.94	&25.07	&40.39	&31.72	&20.89	&26.10 \\
&FeTrIL~\cite{petit2023fetril} &61.41	&48.61	& \color{red} 18.88	& \color{red} 16.14	&53.32	&43.57	&14.69	& \textbf{13.64} \\
&PASS~\cite{zhu2021prototype} &60.33	&51.94	&23.66	&18.78	&45.91	&40.15	&18.00	&  \color{red} 16.69 \\
&PASS w/o Aug &59.02	&55.52	&28.11	&29.55	&48.24	&42.71	&24.01	&26.00 \\ \cline{2-10}
&YONO (Ours) & \color{red} 65.30	& \color{red} 57.66	&24.73	&17.60	& \color{red} 57.23 & \color{red} 54.71	&  \color{red} 14.50 &27.33 \\
&YONO+ (Ours) &\textbf{66.64}&\textbf{60.98} &22.26	& 16.87	&\textbf{58.55} &\textbf{58.42} &\textbf{14.08} & 22.66 \\
\bottomrule
\end{tabular}
\end{adjustbox}
\end{table}

\subsection{Evaluation on Quality of Synthetic Data}
\begin{figure*}[!t]
\centering
\subfigure[CIFAR-100-R]{
\includegraphics[width=0.235\linewidth]{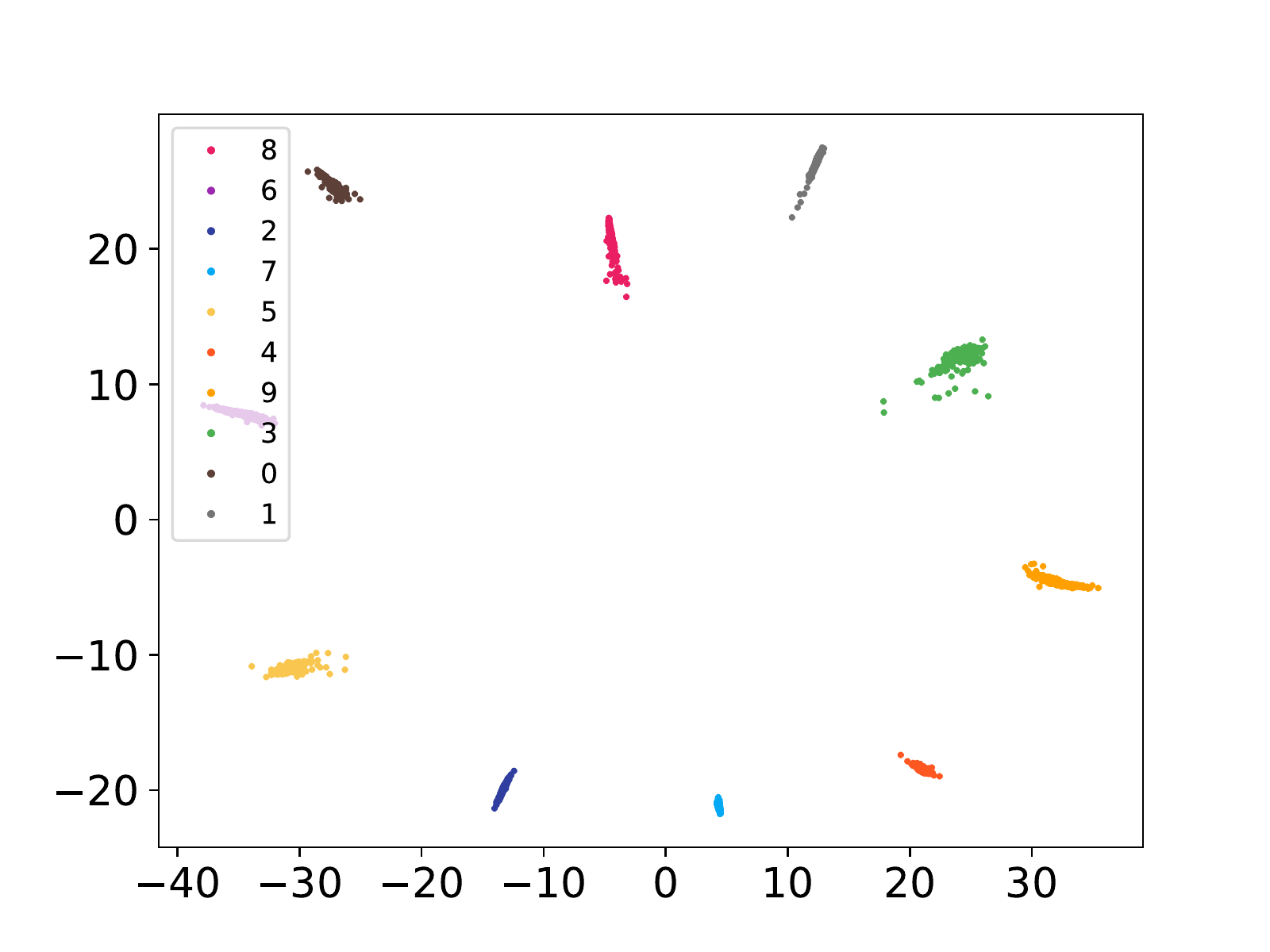}}\hfill
\label{fig:cifar-100-recover}
\subfigure[CIFAR-100-E]{
\includegraphics[width=0.235\linewidth]{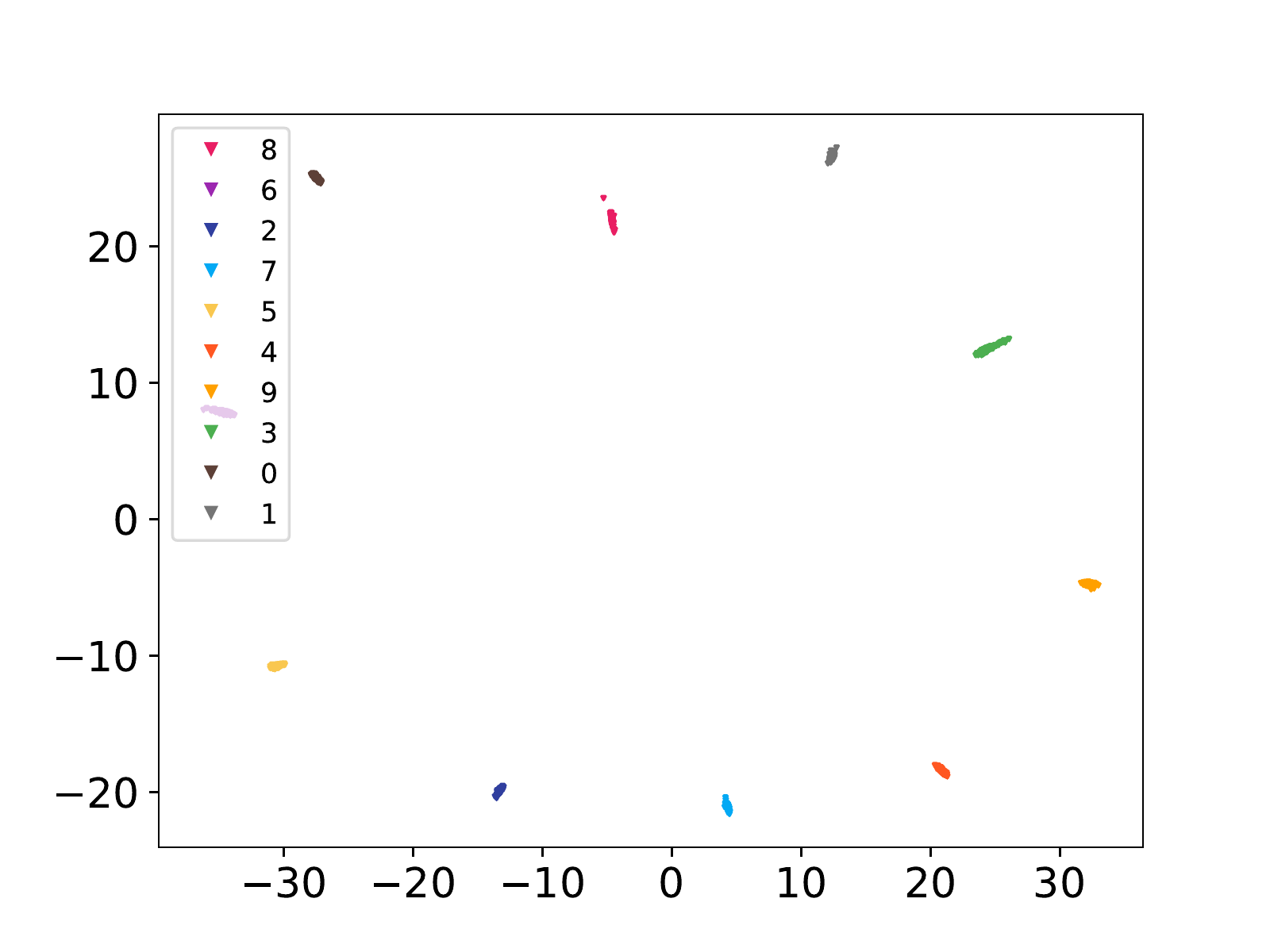}}\hfill
\label{fig:cifar-100-extract}
\subfigure[TinyImageNet-R]{
\includegraphics[width=0.235\linewidth]{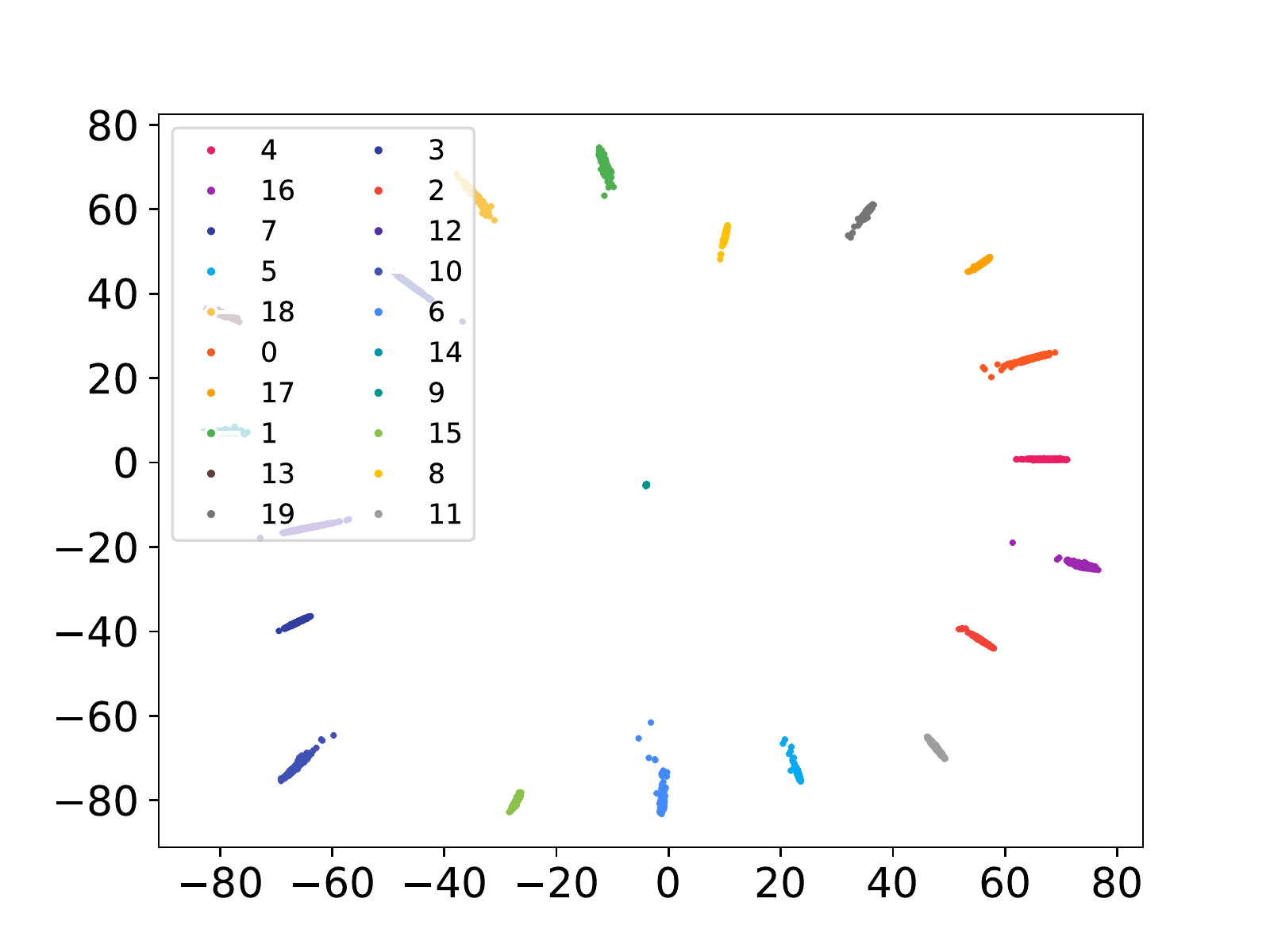}}\hfill
\label{fig:tiny-recover}
\subfigure[TinyImageNet-E]{
\includegraphics[width=0.235\linewidth]{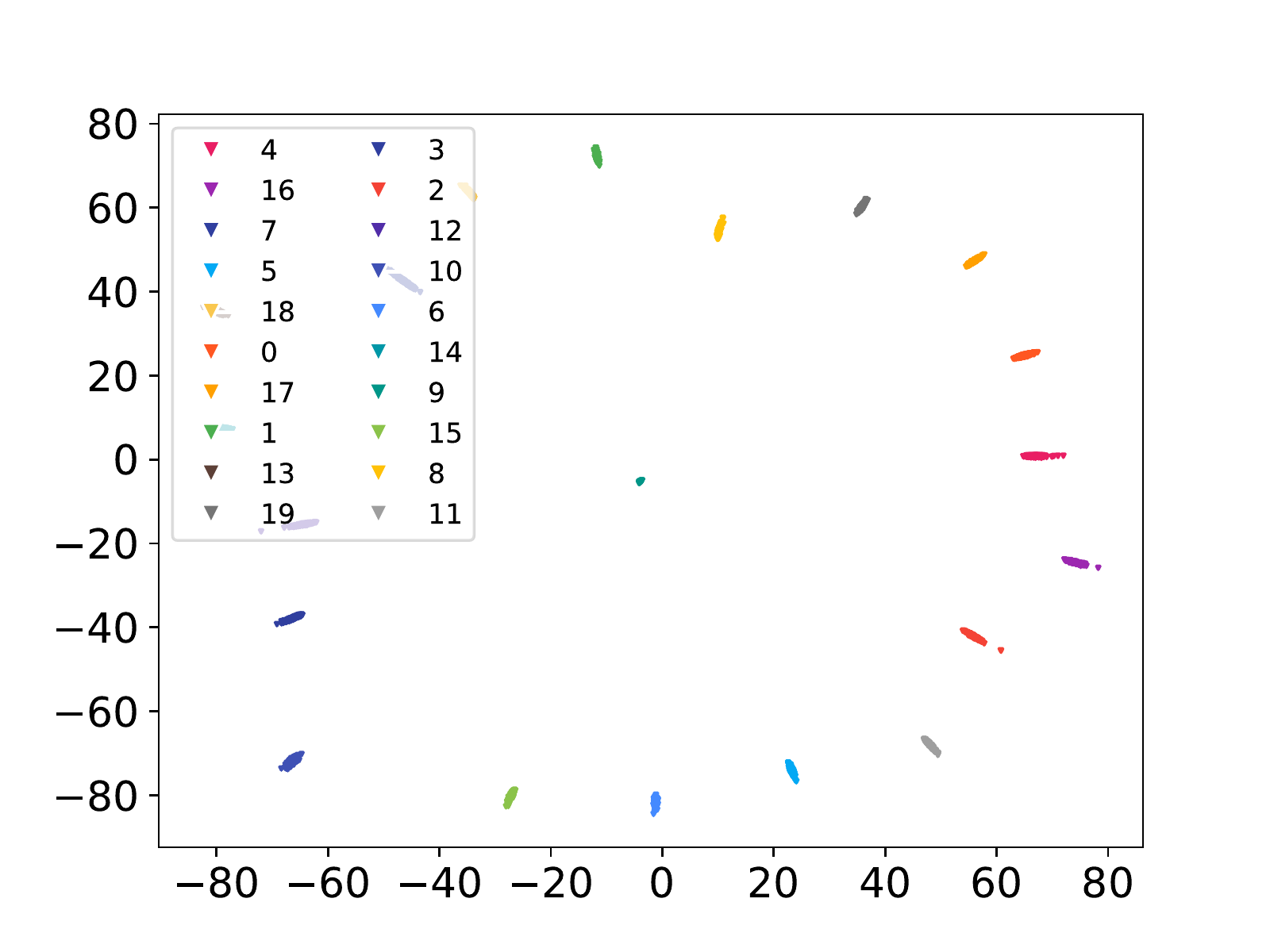}}\hfill
\label{fig:tiny-extract}
\vspace{-0.1in}
\caption{Visualization of the distribution of representations generated from stored prototypes and those encoded by the extractor in the representation space. ``-R" means generated representations, ``-E" means extracted representations.}
\label{fig:recover_provide}
\vspace{-0.2in}
\end{figure*}

Next, we evaluate the quality of synthetic data generated from stored prototypes in YONO+. In this experiment, we randomly choose task $t$ from CIFAR-100 and TinyImageNet when training with 10 phases under zero-base setting. Then we compare the synthetic data for each class from stored prototypes and the representations encoded by the extractor $F(\cdot;\theta)$ after the training of task $t$. Following prior works~\cite{wen2016discriminative,deng2019arcface}, we use two-layer MLP to map the high-dimensional representations into a 2D space for visualization, as shown in Fig.~\ref{fig:recover_provide}. We can observe that the synthetic data generated from stored prototypes in Fig.~\ref{fig:recover_provide} (a) and (c) form a compact cluster, whose distributions are very similar to those of the representations in (b) and (d) encoded by the extractor $F(\cdot;\theta)$. Hence, we can conclude that YONO+ can generate high-quality synthetic data to improve the model performance. 

\subsection{Evaluation on Reliability of YONO+}
\begin{figure*}[!th]
\centering
\subfigure[YONO+ for Task 1]{
\includegraphics[width=0.32\linewidth]{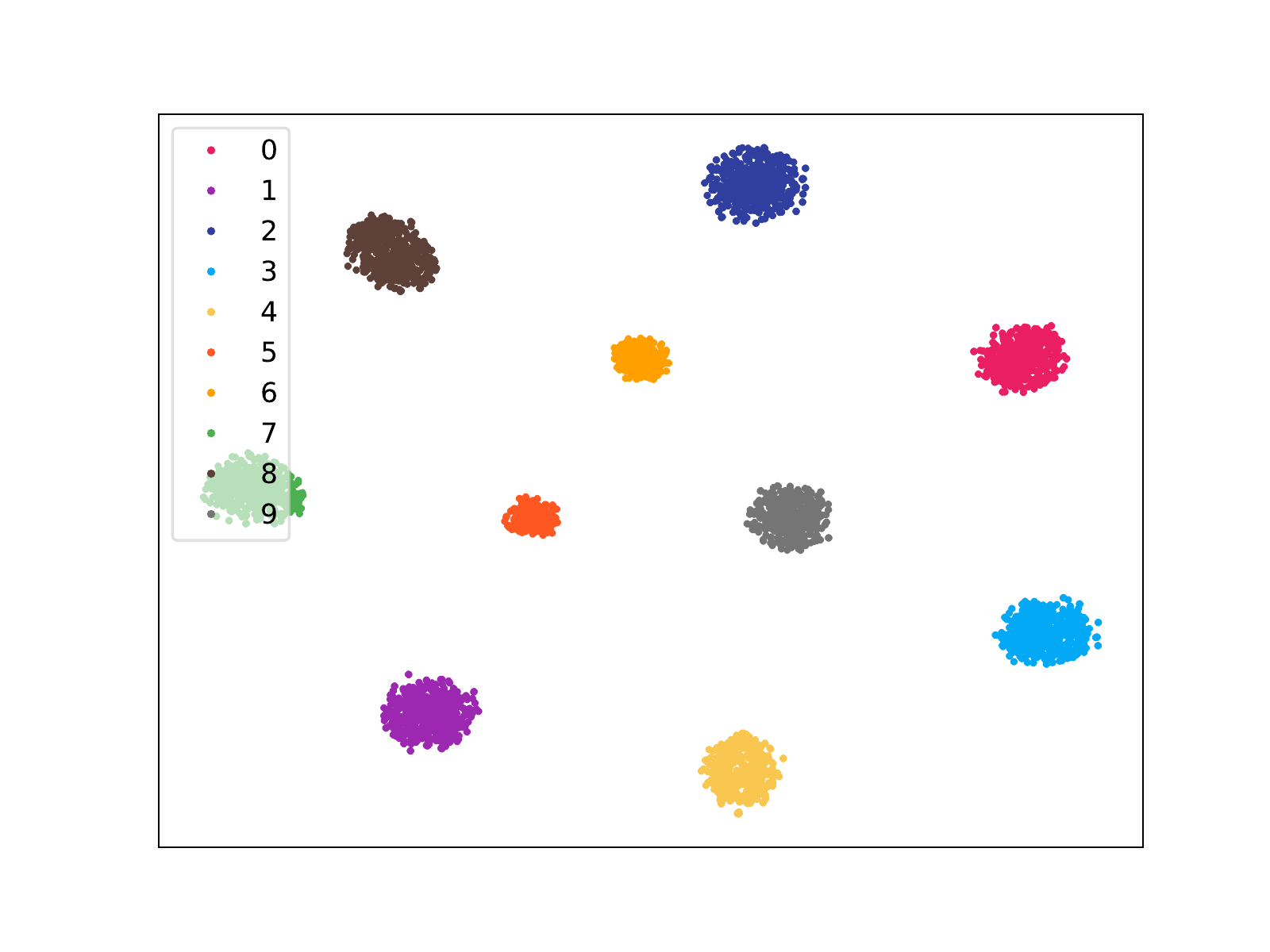}}\label{fig:yono_task1}
\subfigure[YONO+ for Task 2]{
\includegraphics[width=0.32\linewidth]{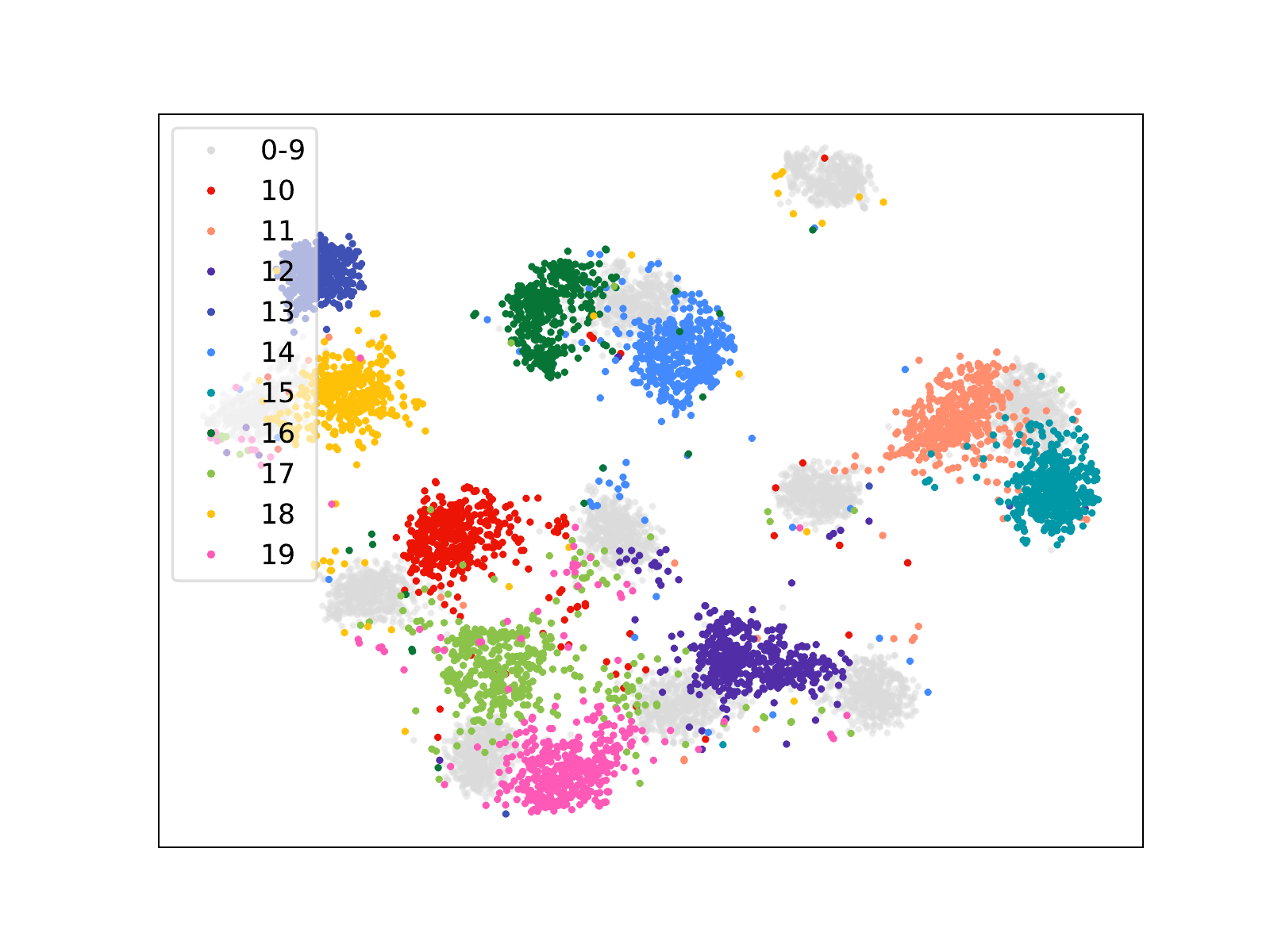}}\label{fig:yono_task2}
\subfigure[YONO+ for Task 3]{
\includegraphics[width=0.32\linewidth]{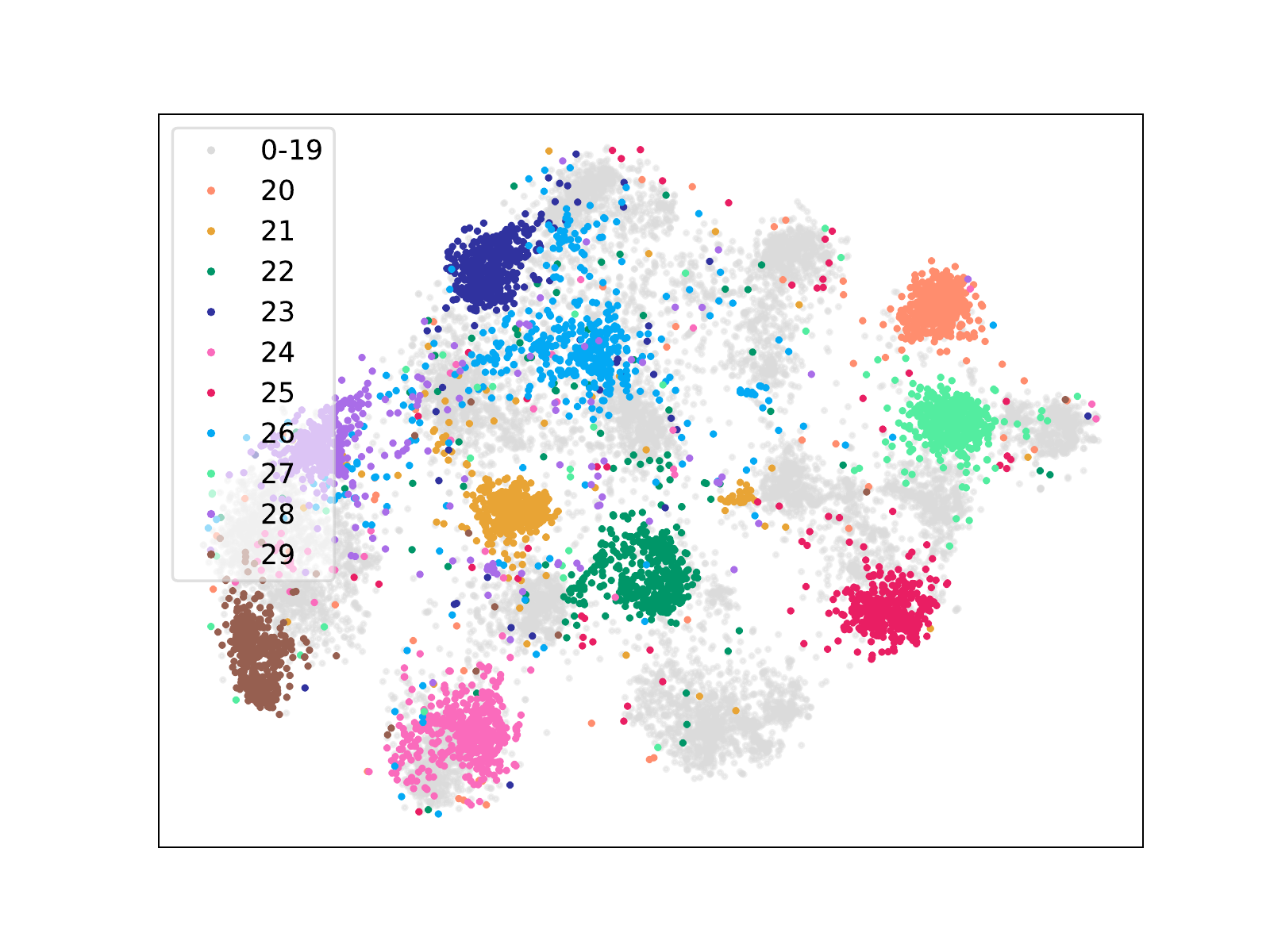}}\label{fig:yono_task3}
\subfigure[PASS for Task 1]{
\includegraphics[width=0.32\linewidth]{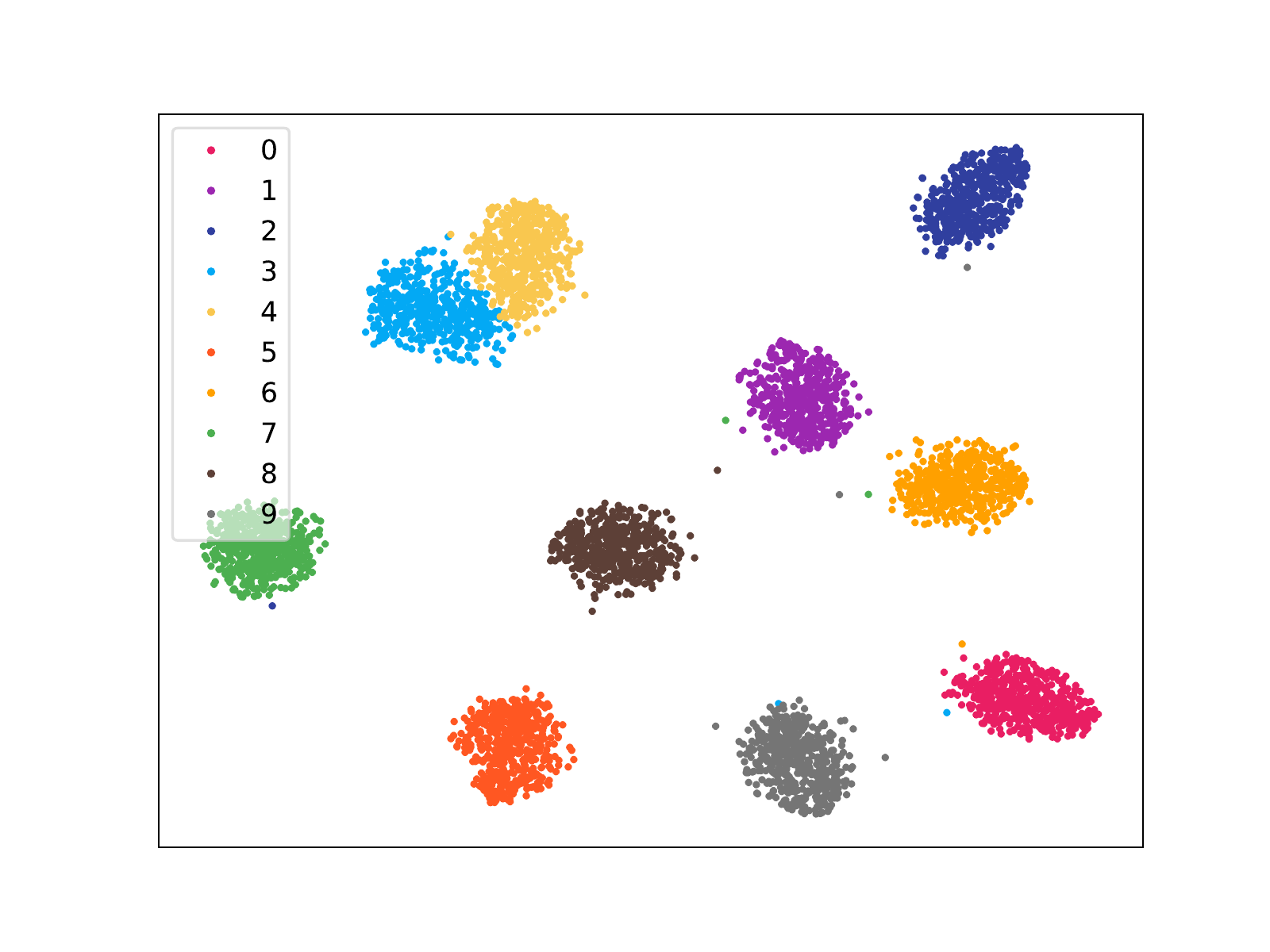}}\label{fig:pass_task1}
\subfigure[PASS for Task 2]{
\includegraphics[width=0.32\linewidth]{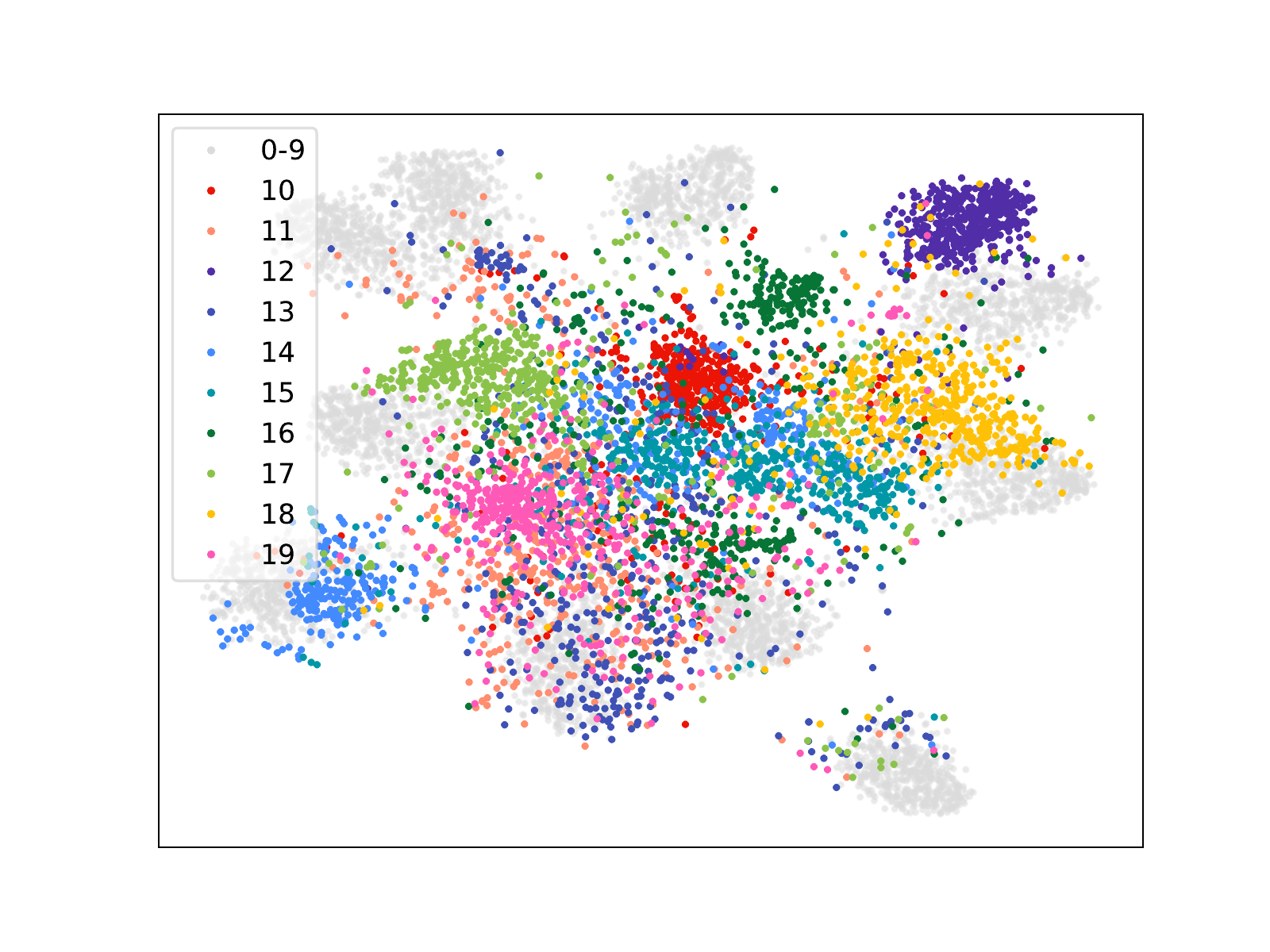}}\label{fig:pass_task2}
\subfigure[PASS for Task 3]{
\includegraphics[width=0.32\linewidth]{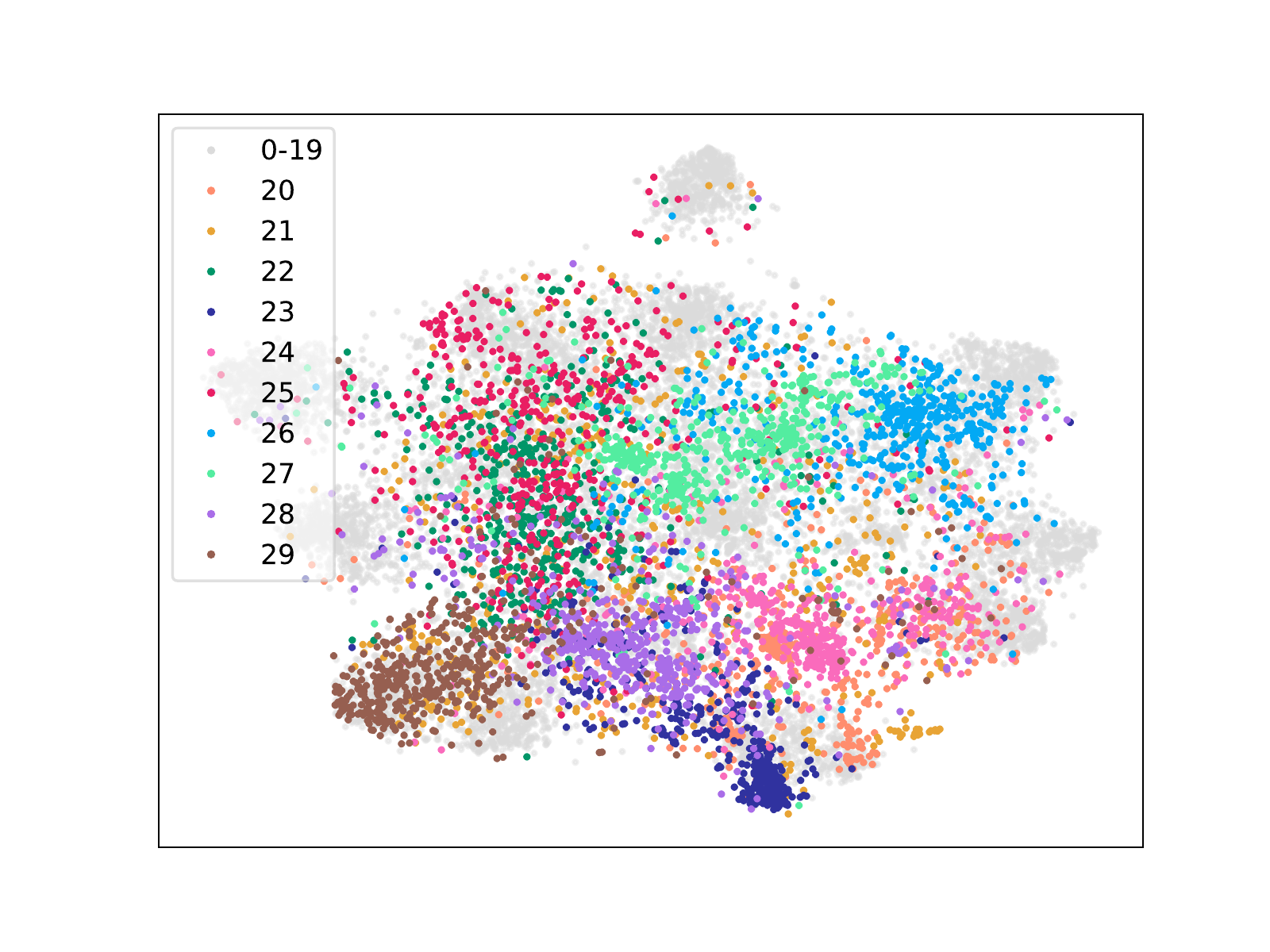}}\label{fig:pass_task3}
\caption{Visualization of the distribution of representations encoded by YONO+ and PASS on CIFAR-100 base-0 phase 10 setting. The lighter gray points in ``Task 2" and ``Task 3" represent the distribution of the previous tasks' data.}
\label{fig:featureMap}
\end{figure*}

In addition, we evaluate the reliability for the proposed YONO+ and PASS in learning the representations from input data samples. Fig.~\ref{fig:featureMap} illustrates the distribution of representations encoded by the extractor on the CIFAR-100 under base-0 phase 10 setting for the first three tasks. Both YONO+ and PASS demonstrate effective maintenance of the decision boundary in the first task. However, in the subsequent tasks 2 and 3, YONO+ can still encode the input data into a representation within a certain boundary, whereas PASS cannot. In Fig.~\ref{fig:featureMap} (b),(c),(e) and (f), the light grey points represent the distribution of data from old tasks. We can see from (b) and (c) that our approach can still separate the old tasks from the current task, while PASS fails to distinguish between the distributions of data from previous tasks and the current one. This is because our attentional mean shift method can form a compact cluster for the samples in each class and also use synthetic data replay to constrain the boundary of old tasks.


\subsection{Ablation Studies}
Finally, we implement ablation studies to explore the impact of some hyper-parameters and components on the performance of our method. 

\begin{wrapfigure}[14]{r}{0.5\textwidth}
\vspace{-0.12in}
  \begin{center}
    \includegraphics[width=0.5\textwidth]{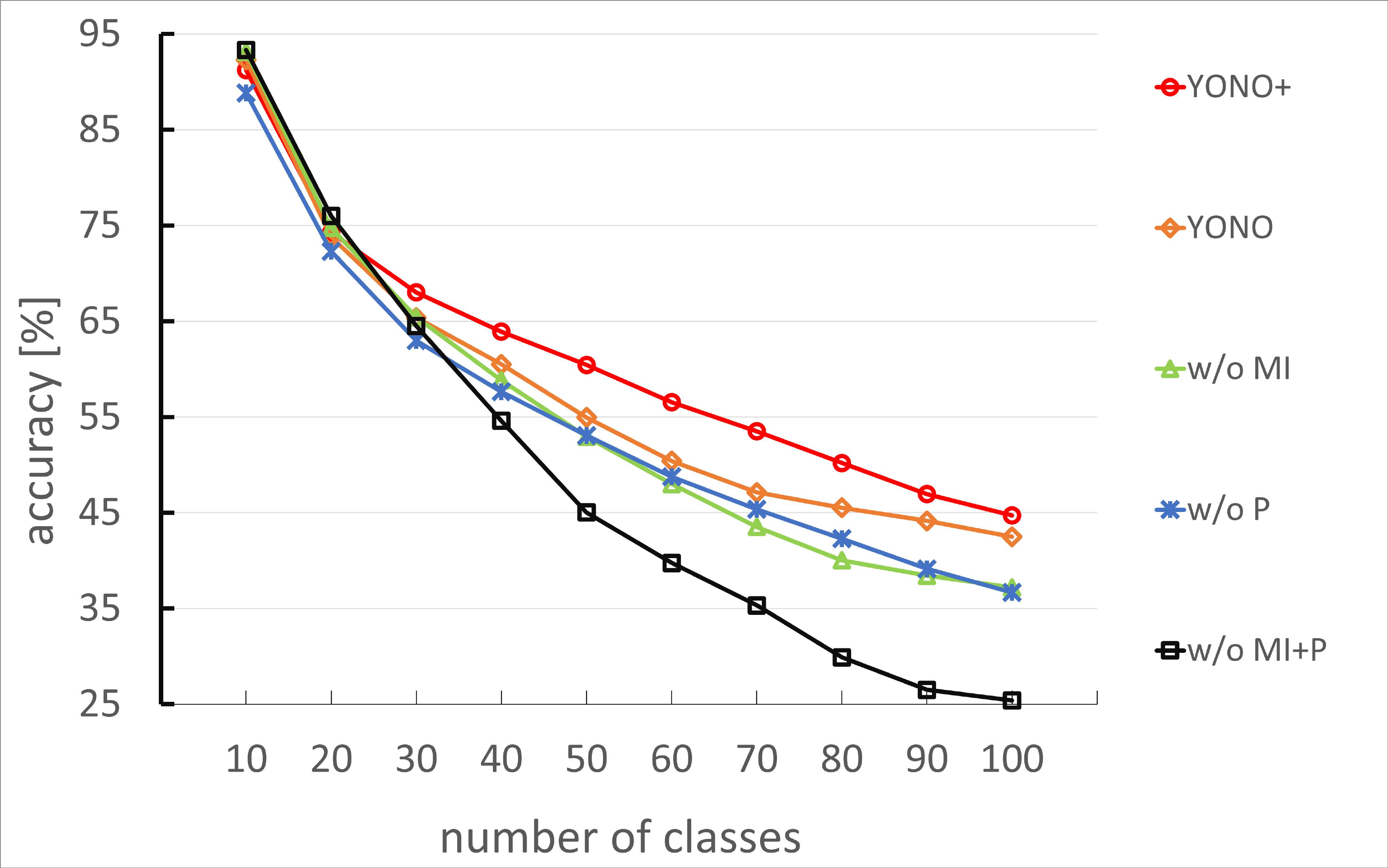}
  \end{center}
  \vspace{-0.2in}
  \caption{Ablation study of different components in YONO+. ``w/o P" means without prototype, ``w/o MI" means without Model Interpolation.}
  \label{fig:ablation}
\end{wrapfigure}

\noindent
\textbf{Effect of important components.} 
We first study the impact of some important components, such as prototype, synthetic data replay, model interpolation (MI), on prediction accuracy. Fig.~\ref{fig:ablation} illustrates the comparison results on CIFAR-100 under 10 phases setting. We can observe from it that the saved prototype plays an important role in our proposed methods. The prediction accuracy will drop a lot without it. Additionally, YONO+ with synthetic data replay can slightly improve the prediction accuracy compared to YONO. Besides, MI can help improve model performance since it can retain the knowledge of prior model while ensuring the good performance of the current model.
\noindent
\textbf{Effect of important hyperparameters.} We also explore the impact of some hyperparameters, such as margin penalty $\delta$ in Arcface (Eq.~\eqref{equ:arcface}), $\lambda$ in attentional mean-shift method, $\beta$ for model interpolation in Eq.~\eqref{eq:theta_update} and $\eta'$ for “partial freezing” of classifier in Eq.~\eqref{eq:eta_classifier}, on the prediction accuracy of our method. The detailed experimental results are presented in Appendix~\ref{app:ablation}.

%% file: relatedwork.tex
\section{Related Work}\label{sec:relatedwork}

\noindent \textbf{Regularization-based method}. It aims to alleviate catastrophic forgetting by introducing additional regularization terms to correct the gradients and protect the old knowledge learned by the model~\cite{li2017learning, rannen2017encoder,kirkpatrick2017overcoming,hou2019learning}. For example, some works~\cite{roady2020stream,schwarz2018progress,wang2021ordisco} adopt weights regularization to regularize the variation of model parameters. However, it is very hard to design reasonable and reliable metrics to measure the importance of model parameters. Others~\cite{li2017learning,pham2021dualnet,douillard2020podnet} mainly adopt KD to transfer the output of prediction function from previously-learned tasks to the current task. However, these methods need to replay some old data samples. 

\noindent \textbf{Replay-based method}. The replay-based methods mainly include experience replay~\cite{riemer2018learning,lopez2017gradient} and generative replay~\cite{wu2018memory,cong2020gan,rostami2019complementary}. The former approach stores some old training examples within a memory buffer while the latter mainly uses generative models to generate data samples of old classes. For experience replay, some approaches mainly adopt rehearsal samples with knowledge distillation~\cite{wu2019large,rebuffi2017icarl,douillard2020podnet} and some~\cite{riemer2018learning,chaudhry2018efficient} apply regularization on the gradients for the sake of using the rehearsal samples more sufficiently. While these methods can mitigate catastrophic forgetting, they will suffer from data privacy issues and need a large-size memory. In order to mitigate privacy issues, some works~\cite{lesort2019generative,rios2018closed} adopt deep generative models to generate pseudo samples of previous tasks. Nevertheless, these methods either suffer from the instability of generative models or still need to store some old examples. 

\noindent \textbf{Architecture-based method}. This approach can be divided into parameters isolation~\cite{serra2018overcoming,xue2022meta} and dynamic architecture. 
Parameters isolation methods adopt individual parameters for each task, thus they need a large memory to store the extended network for each previous task during training~\cite{mallya2018packnet, serra2018overcoming, yoon2017lifelong, yan2021dynamically}. In addition, the architecture-based methods~\cite{li2021continual,ostapenko2019learning} dynamically expand the network if its capacity is not large enough for new tasks. Those methods can achieve remarkable performance, but they are not applicable to a large number of tasks. 

\noindent \textbf{Most Relevant Work.} The work closely related to ours is PASS~\cite{zhu2021prototype}, which saves one prototype for each class and then augments the prototype via Gaussian noise for model training. This method can save the storage of memory buffers and maintain the decision boundary of previous tasks to some degree. However, PASS uses the mean value of representations in each class as a prototype, which cannot represent the centroid of a cluster exactly. As a result, its prediction performance degrades after prototype augmentation since some representations of old classes may overlap with other similar classes. In contrast, we try to learn a more representative prototype from data samples in each class so that the class margins can be maximized, thereby mitigating forgetting.


%% file: conclude.tex
\section{Conclusion}
In this work, we developed two non-exemplar-based methods, YONO and YONO+, for class-incremental learning. Specifically, YONO only needs to store and replay one prototype for each class without generating synthetic data from stored prototypes. As an extension of YONO, YONO+ proposed to create synthetic replay data from stored prototypes via a high-dimensional rotation matrix and Gaussian noise. The evaluation results on multiple benchmarks demonstrated that both YONO and YONO+ can significantly outperform the baselines in terms of accuracy and average forgetting. In particular, the proposed YONO achieved comparable performance to YONO+ with data synthesis. Importantly, this work offered a new perspective of optimizing class prototypes for exemplar-free incremental learning.

%% file: appendix.tex
\section{More Ablation Studies}\label{app:ablation}
We conduct more ablation studies to explore some important hyper-parameters on model performance. 
\subsection{Effect of $\lambda$ in mean-shift method}
Table.~\ref{table:lambda_ablation} shows the impact of hyper-parameter $\lambda$ in mean-shift on prototype learning using CIFAR100. We can observe from it that as $\lambda$ increases from 0.3 to 0.9, the prediction accuracy of YONO+ almost remains the same. Thus, we can conclude that the proposed method is not sensitive to $\lambda$ that controls the steps in the attentional mean-shift algorithm. We choose $\lambda=0.6$ in our experiments.
\begin{table}[!htb]
\caption{Effect of $\lambda$ on the accuracy of different tasks on CIFAR-100 with base-0 10 phases setting.}
\label{table:lambda_ablation}
\centering
\begin{adjustbox}{width=0.9\textwidth}
\begin{tabular}{c|c|c|c|c|c|c|c|c|c|c}
\toprule
\multicolumn{11}{c}{Base-0-10 phases Accuracy on CIFAR-100 with different $\lambda$}      \\ \hline
$\lambda$ &Task 1 &Taks 2 &Task 3 &Task 4 &Task 5 &Task 6 &Task 7 &Task 8 &Task 9 &Task 10 \\ \hline
0.3	&0.925	&0.6855	&0.6427	&0.5830	&0.5840	&0.5425	&0.5279	&0.4959	&0.4701	&0.4457 \\
0.4	&0.9136	&0.7433	&0.6813 &0.6396 &0.6037	&0.5652	&0.5335 &0.5003 &0.4690 &0.4476 \\
0.5	&0.926	&0.6900	&0.6450	&0.5936	&0.5789	&0.5376	&0.5189	&0.4862	&0.4628	&0.4333 \\
0.6	&0.9134	&0.7430	&0.6810	&0.6417	&0.6055	&0.5667	&0.5359	&0.5022	&0.4706	&0.4478 \\
0.7	&0.9128	&0.7455	&0.6851	&0.6422	&0.6049	&0.5663	&0.5365	&0.5026	&0.4691	&0.4473 \\
0.8	&0.9140	&0.7459	&0.6825	&0.6426	&0.6059	&0.5675	&0.5370	&0.5024	&0.4706	&0.4481 \\
0.9	&0.9135	&0.7463	&0.6823	&0.6426	&0.6063	&0.5682	&0.5362	&0.5025	&0.4689	&0.4474 \\
\bottomrule
\end{tabular}
\end{adjustbox}
\end{table}
\subsection{Effect of margin penalty $\delta$.}
\begin{figure}[!thb]
\centering
\subfigure[Ablation-$\delta$]{
\includegraphics[width=0.32\linewidth]{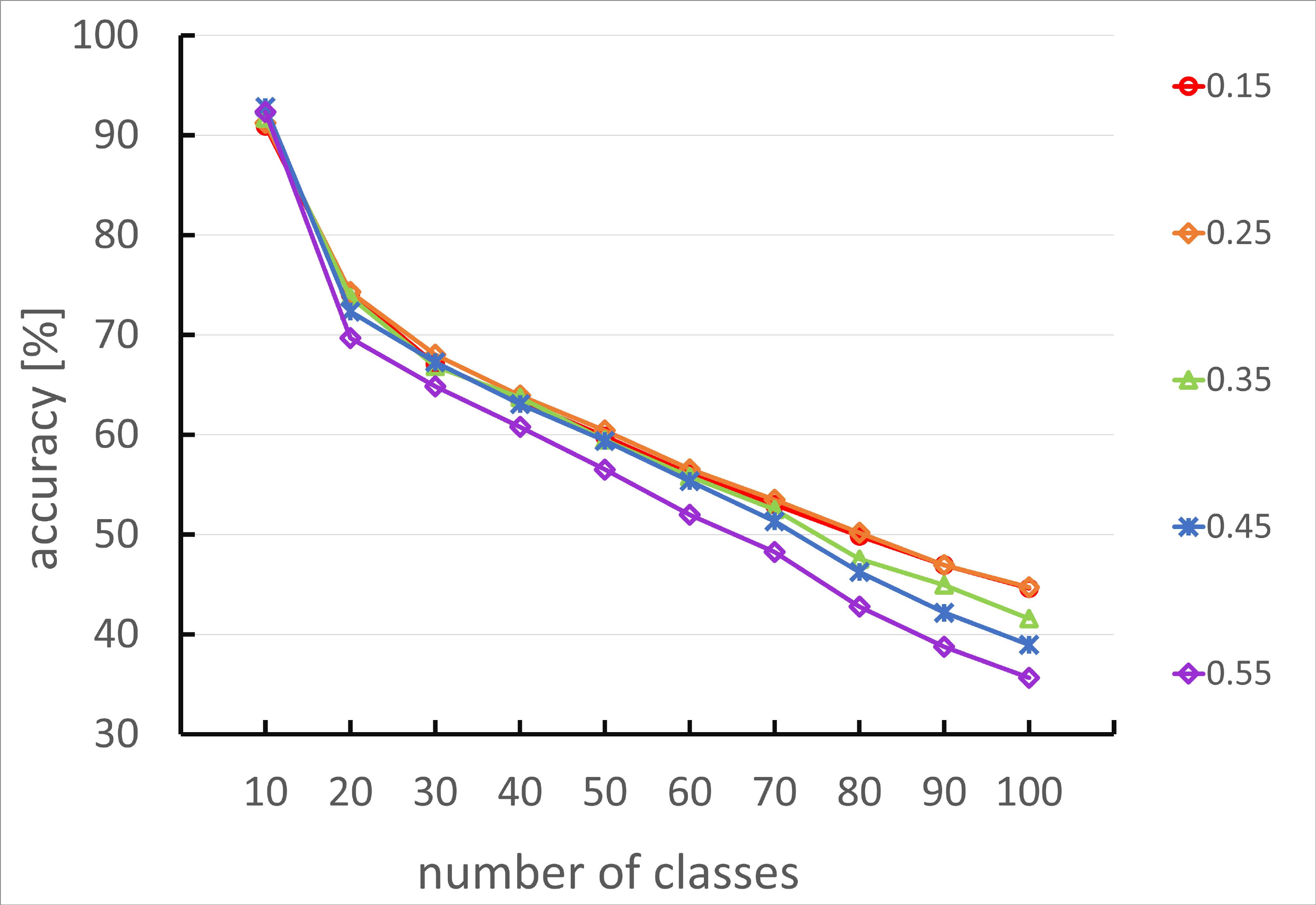}}
\label{fig:ablation-margin}
\subfigure[Ablation-$\beta$]{
\includegraphics[width=0.32\linewidth]{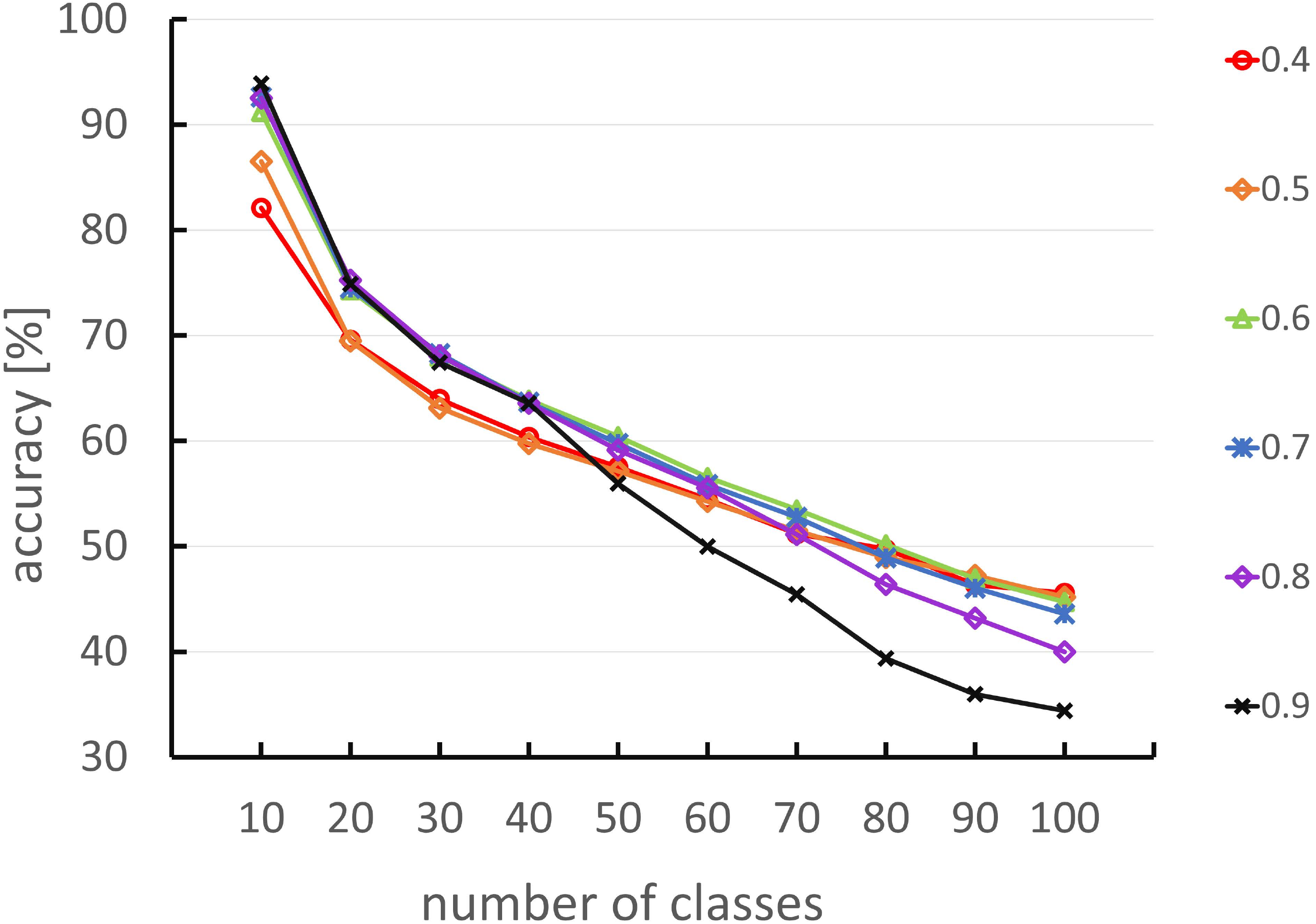}}
\label{fig:ablation-beta}
\subfigure[Ablation-$\eta'$]{
\includegraphics[width=0.32\linewidth]{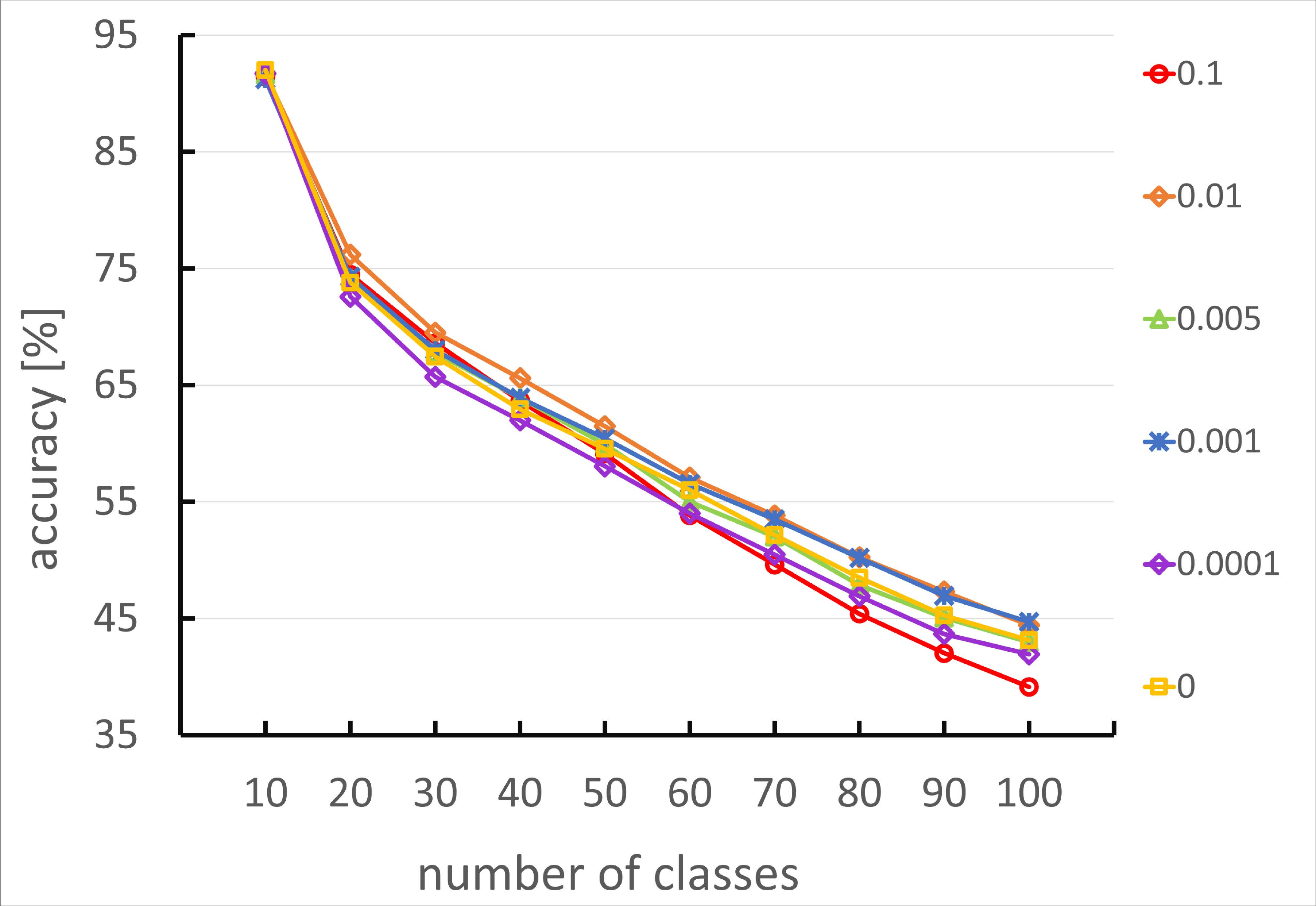}}
\label{fig:ablation-tiny}
\vspace{-0.1in}
\caption{The ablation study of $\delta$, $\beta$ and $\eta'$ on the prediction results.}
\label{fig:ablation-margin-tiny}
\vspace{-0.2in}
\end{figure}

We also investigate the effect of margin penalty $\delta$ in the Arcface on the prediction accuracy of our method. It can be observed from Fig.~\ref{fig:ablation-margin-tiny} (a) that as $\delta$ increases from $0.15$ to $0.45$, the average accuracy drops after $\delta > 0.25$. Therefore, $\delta$ should be limited to no more than 0.25. According to this, we choose $\delta=0.25$ in our experiments.


\subsection{Effect of parameter $\beta$ in model interpolation}
In addition, we investigate the influence of hyperparameter $\beta$ in model interpolation on the prediction accuracy, as shown in Fig.~\ref{fig:ablation-margin-tiny} (b). It can be seen that when $\beta=0.6$, the proposed method has the best performance. When $\beta$ increases from 0.6 to 0.9, the prediction accuracy gradually drops.  

\subsection{Effect of small learning rate $\eta'$}
What is more, we study the effect of learning rate $\eta'$ in the ``Partial Freezing'' of the classifier on the model performance. As illustrated in Fig.~\ref{fig:ablation-margin-tiny} (c), it can be observed that when $\eta'$ is 0 or larger than $0.005$, the prediction accuracy will drop significantly. As $\eta'$ is a very small value, such as 0.001, the proposed method has the best performance. Hence, we choose $\eta'=0.001$ in our experiment.

\section{Evaluation on Half-base Setting}
\begin{figure}[!thb]
\centering
\subfigure[CIFAR-half-base-phases5]{
\includegraphics[width=0.46\linewidth]{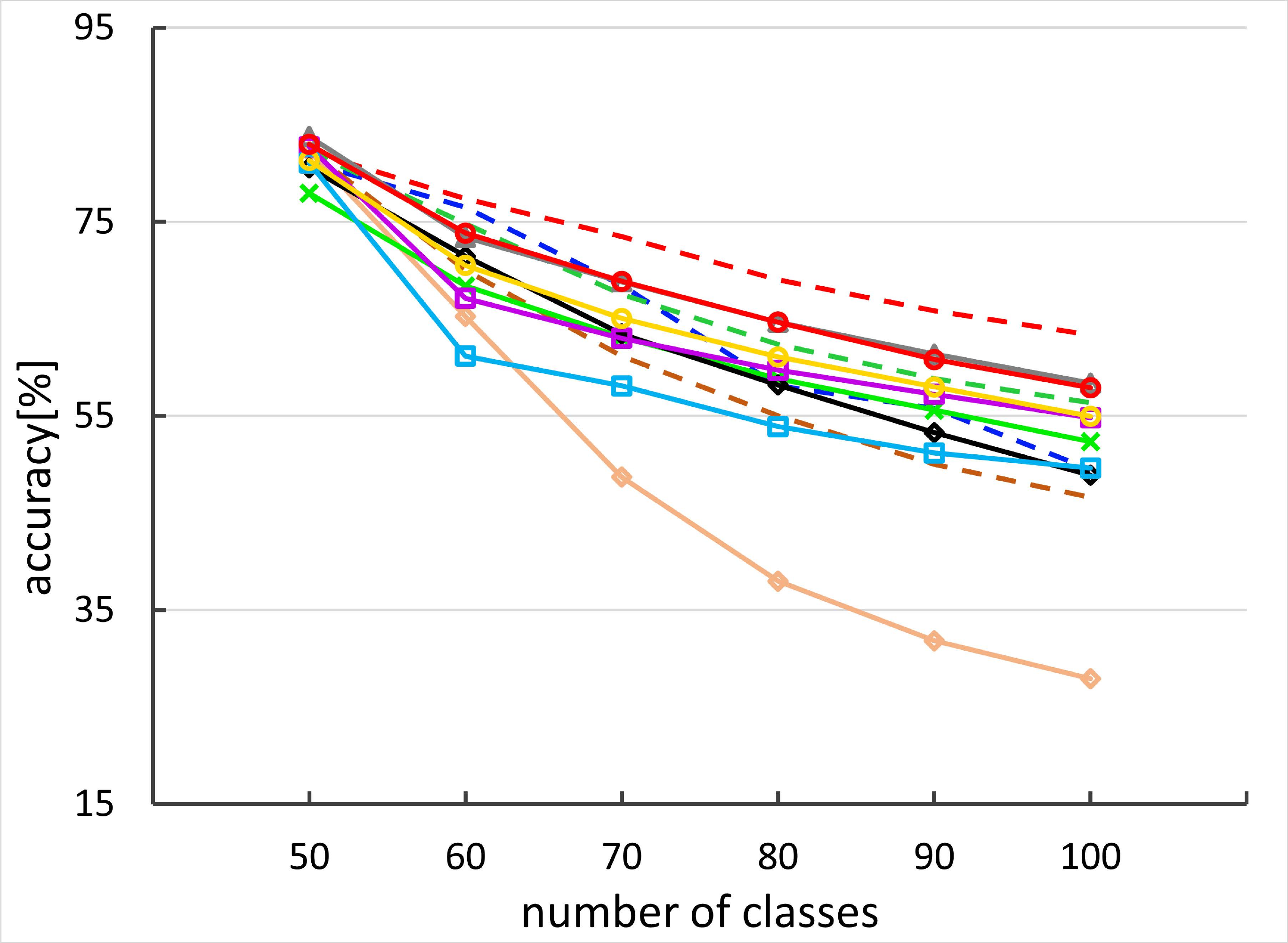}}
\label{fig:cifar-hb-5}
\subfigure[CIFAR-half-base-phases10]{
\includegraphics[width=0.52\linewidth]{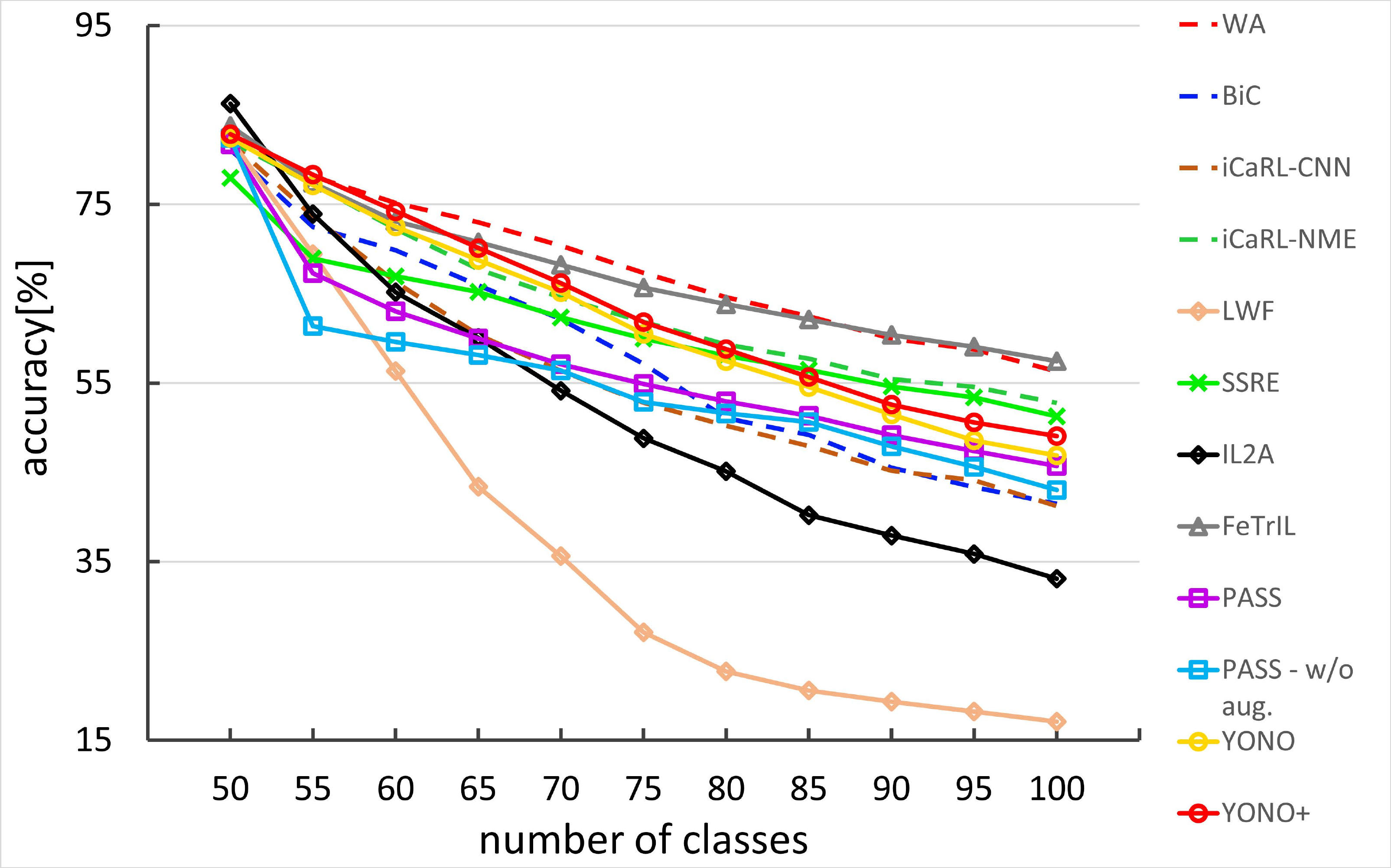}}
\label{fig:cifar-hb-10}
\subfigure[Tiny-half-base-phases5]{
\includegraphics[width=0.46\linewidth]{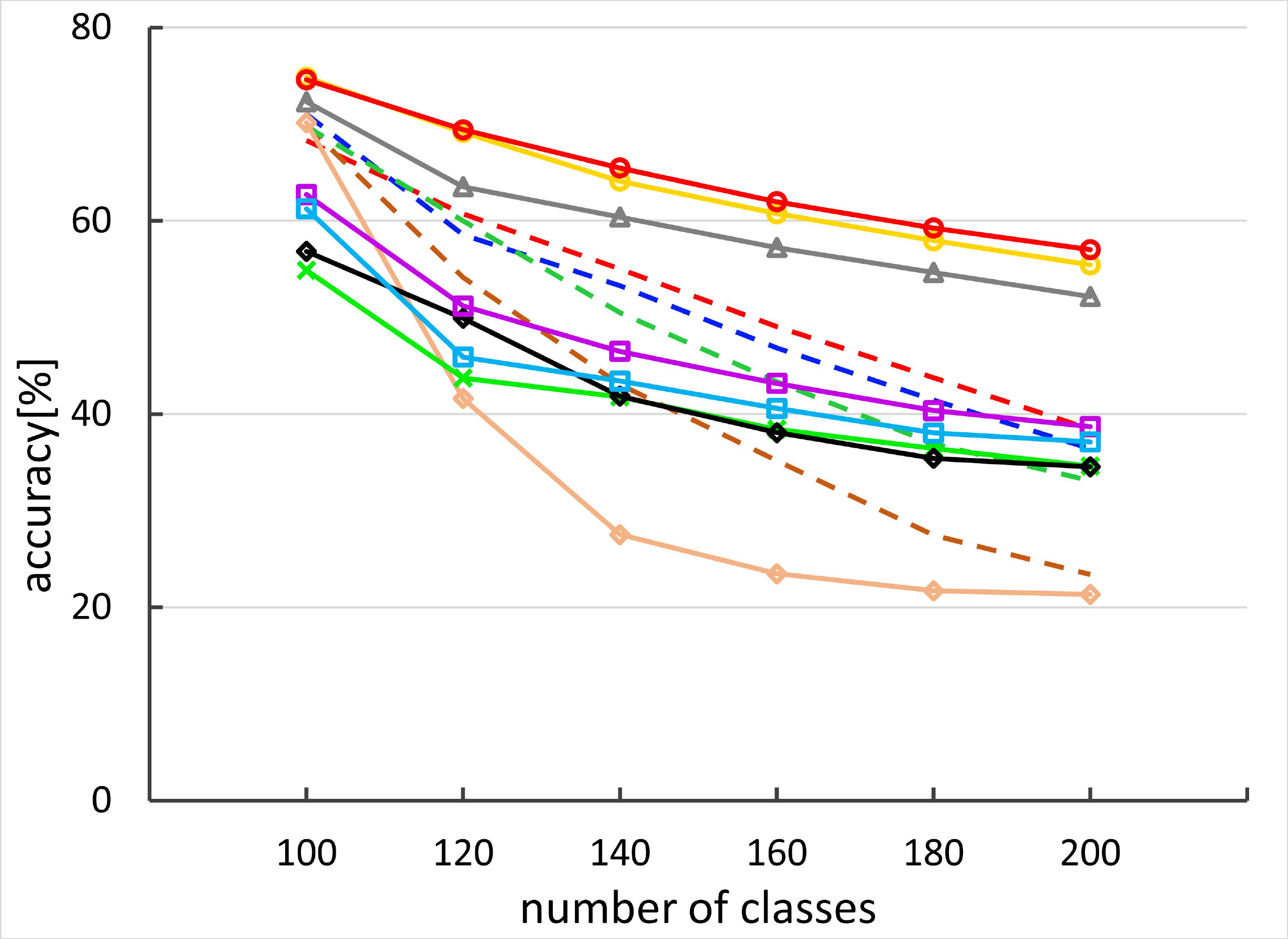}}
\label{fig:tiny-hb-5}
\subfigure[Tiny-half-base-phases10]{
\includegraphics[width=0.52\linewidth]{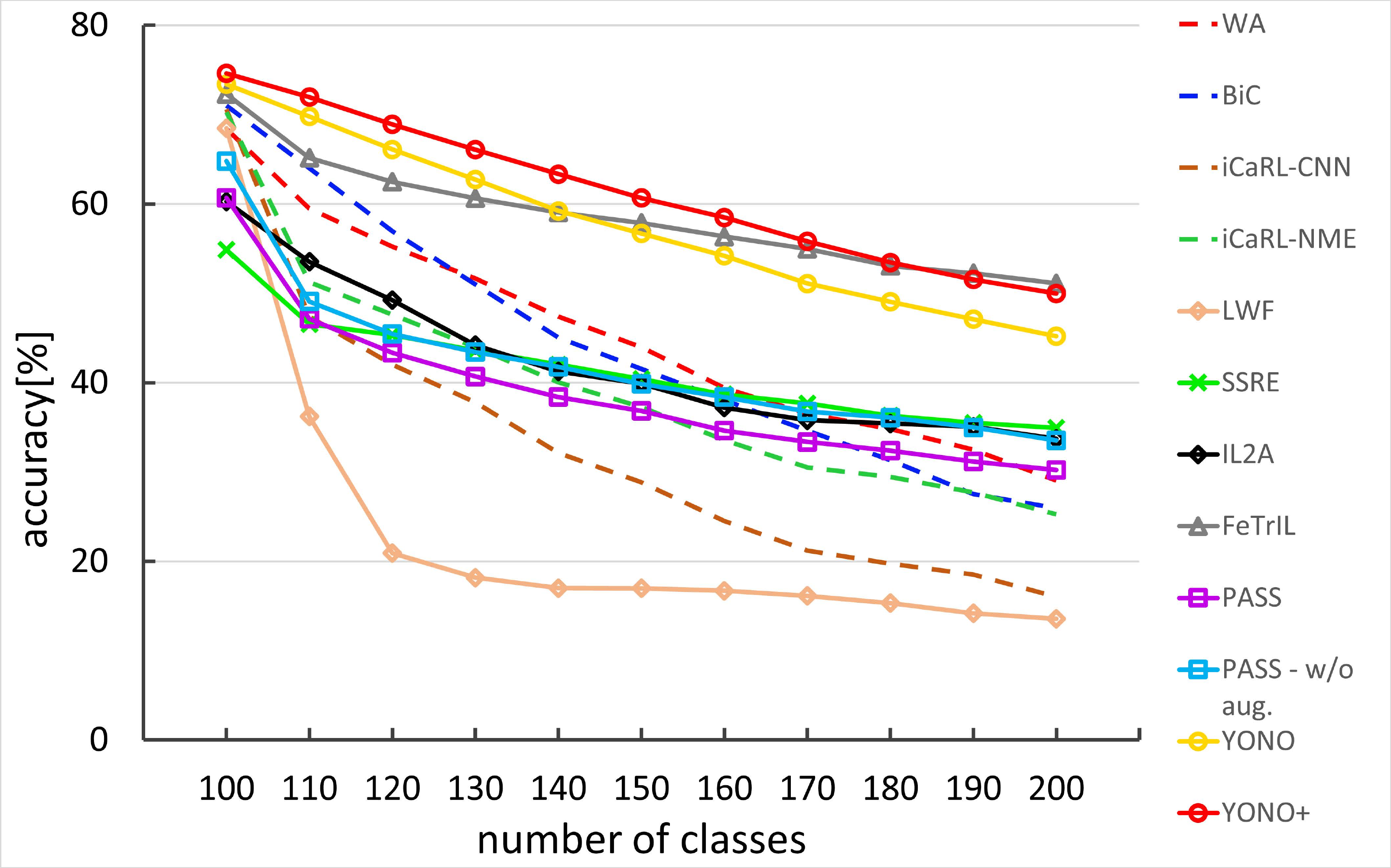}}
\label{fig:tiny-hb-10}
\vspace{-0.15in}
\caption{Accuracy comparison of different methods on CIFAR-100 and TinyImageNet under different settings using 3 random seeds. Solid lines represent non-exemplar-based approaches while the dashed lines denote exemplar-based methods.}
\label{fig:sota_half_cifar_tiny}
\vspace{-0.1in}
\end{figure}

We also compare the proposed methods with the baselines under half-base setting with 5 and 10 phrases. Fig.~\ref{fig:sota_half_cifar_tiny} illustrates the accuracy comparison of different approaches on CIFAR-100 and TinyImageNet using three random seeds. We can observe that the proposed YONO and YONO+ outperform the non-exemplar-based methods on both CIFAR100 and TinyImageNet under different settings, except for one special scenario with 10 phrases where FeTril has slightly higher accuracy than our methods on CIFAR-100. In addition, both YONO and YONO+ can achieve higher accuracy than most exemplar-based methods on TinyImageNet.


